\documentclass[10pt,twocolumn,letterpaper]{article}
\PassOptionsToPackage{table}{xcolor}
\usepackage{cvpr}

\usepackage{tabularx}
\usepackage{array}
\usepackage{multirow}
\usepackage{cuted}
\usepackage{adjustbox}
\usepackage{tikz}
\definecolor{bestgreen}{HTML}{C6E0B4}
\definecolor{secondgreen}{HTML}{E2EFDA}
\newcommand{\best}[1]{\cellcolor{bestgreen}\textbf{#1}}
\newcommand{\secondbest}[1]{\cellcolor{secondgreen}#1}
\definecolor{cvprblue}{rgb}{0.21,0.49,0.74}
\usepackage[breaklinks,colorlinks,citecolor=cvprblue,
            linkcolor=cvprblue,urlcolor=cvprblue]{hyperref}

\newcommand{\pithree}{$\pi^3$}
\newcommand{\pilong}{$\pi^3$-Long}
\newcommand{\dastream}{DA3-Streaming}
\newcommand{\figref}[1]{Fig.~\ref{#1}}
\newcommand{\secref}[1]{Sec.~\ref{#1}}
\newcommand{\tabref}[1]{Tab.~\ref{#1}}
\newcommand{\tabsref}[1]{Tabs.~\ref{#1}}

\newcommand{\ours}{FFVO}
\newcommand{\ourspost}{\ours~(Ours)}
\definecolor{improvegreen}{HTML}{2E7D32}

\newcommand{\greenperc}[1]{\,\textcolor{improvegreen}{\scriptsize{(+#1\%)}}}

\def\confName{3DV\xspace}
\def\confYear{2027\xspace}

\title{\ours: A Feedforward Pose Decoder for Long-Horizon Visual Odometry}

\author{
Meng-Li Shih$^{1,2}\thanks{:This work was done during Meng-Li's internship at Waymo LLC.}$,~ Shih-Yang Su$^{1}$, Yuliang Zou$^{1}$, Hao Xiang$^{1}$, Haidong Zhu$^{1}$,\\
Vincent Casser$^{1}$, Brian Curless$^{2,3}$, Dmitry Kalenichenko$^{1}$, Mingxing Tan$^{1}$, Dragomir Anguelov$^{1}$\\\\
 $^{1}$Waymo LLC \qquad $^{2}$University of Washington \qquad $^{3}$Google DeepMind\\ 
{\tt\small mlshih@cs.washington.edu}\\
}
\begin{document}

\maketitle

\newcommand{\kittireconpath}{images/kitti_cloud_plot/cropped/}
\newcommand{\kittireconrowtwoheight}{0.434375\linewidth}
\newcommand{\kittireconimg}[2]{%
  \begin{tikzpicture}[inner sep=0,outer sep=0]
    \node[anchor=south west,inner sep=0] (img) at (0,0) {%
      \includegraphics[width=\linewidth,keepaspectratio,trim=#2,clip]{\kittireconpath#1}%
    };
  \end{tikzpicture}%
}
\newcommand{\kittireconimgpaddedtop}[3]{%
  \begin{tikzpicture}[inner sep=0,outer sep=0]
    \node[
      anchor=south west,
      inner sep=0,
      outer sep=0,
      minimum width=\linewidth,
      minimum height=#3,
      fill=white
    ] (frame) at (0,0) {};
    \node[anchor=south west,inner sep=0,outer sep=0] at (frame.south west) {%
      \includegraphics[width=\linewidth,keepaspectratio,trim=#2,clip]{\kittireconpath#1}%
    };
  \end{tikzpicture}%
}
\newcommand{\kittireconimgpaddedtopwithinsettr}[4]{%
  \begin{tikzpicture}[inner sep=0,outer sep=0]
    \node[
      anchor=south west,
      inner sep=0,
      outer sep=0,
      minimum width=\linewidth,
      minimum height=#3,
      fill=white
    ] (frame) at (0,0) {};
    \node[anchor=south west,inner sep=0,outer sep=0] at (frame.south west) {%
      \includegraphics[width=\linewidth,keepaspectratio,trim=#2,clip]{\kittireconpath#1}%
    };
    \node[
      anchor=north east,
      xshift=-1.5pt,
      yshift=-1.5pt,
      inner sep=0.8pt,
      fill=white,
      draw=black,
      line width=0.25pt
    ] at (frame.north east) {%
      \includegraphics[width=0.23\linewidth]{#4}%
    };
  \end{tikzpicture}%
}
\newcommand{\kittireconimgpaddedtopwithinsettrscaled}[5]{%
  \begin{tikzpicture}[inner sep=0,outer sep=0]
    \node[
      anchor=south west,
      inner sep=0,
      outer sep=0,
      minimum width=\linewidth,
      minimum height=#3,
      fill=white
    ] (frame) at (0,0) {};
    \node[anchor=south west,inner sep=0,outer sep=0] at (frame.south west) {%
      \includegraphics[width=\linewidth,keepaspectratio,trim=#2,clip]{\kittireconpath#1}%
    };
    \node[
      anchor=north east,
      xshift=-1.5pt,
      yshift=-1.5pt,
      inner sep=0.8pt,
      fill=white,
      draw=black,
      line width=0.25pt
    ] at (frame.north east) {%
      \includegraphics[width=#5\linewidth]{#4}%
    };
  \end{tikzpicture}%
}
\newcommand{\kittireconimgpaddedtopwithinsettrellipse}[8]{%
  \begin{tikzpicture}[inner sep=0,outer sep=0]
    \node[
      anchor=south west,
      inner sep=0,
      outer sep=0,
      minimum width=\linewidth,
      minimum height=#3,
      fill=white
    ] (frame) at (0,0) {};
    \node[anchor=south west,inner sep=0,outer sep=0] (img) at (frame.south west) {%
      \includegraphics[width=\linewidth,keepaspectratio,trim=#2,clip]{\kittireconpath#1}%
    };
    \begin{scope}[shift={(img.south west)}, x={(img.south east)}, y={(img.north west)}]
      \draw[red, dash pattern=on 2pt off 2pt, line cap=round, line width=0.8pt] (#6,#7) ellipse [#8];
    \end{scope}
    \node[
      anchor=north east,
      xshift=-1.5pt,
      yshift=-1.5pt,
      inner sep=0.8pt,
      fill=white,
      draw=black,
      line width=0.25pt
    ] at (frame.north east) {%
      \includegraphics[width=#5\linewidth]{#4}%
    };
  \end{tikzpicture}%
}
\newcommand{\kittireconimgscaled}[3]{%
  \begin{tikzpicture}[inner sep=0,outer sep=0]
    \node[anchor=south west,inner sep=0] (img) at (0,0) {%
      \includegraphics[width=#3\linewidth,keepaspectratio,trim=#2,clip]{\kittireconpath#1}%
    };
  \end{tikzpicture}%
}
\newcommand{\kittireconimgscaledwithinset}[5]{%
  \begin{tikzpicture}[inner sep=0,outer sep=0]
    \node[anchor=south west,inner sep=0] (img) at (0,0) {%
      \includegraphics[width=#3\linewidth,keepaspectratio,trim=#2,clip]{\kittireconpath#1}%
    };
    \node[
      anchor=north west,
      xshift=1.5pt,
      yshift=-1.5pt,
      inner sep=0.8pt,
      fill=white,
      draw=black,
      line width=0.25pt
    ] at (img.north west) {%
      \includegraphics[width=#5\linewidth]{#4}%
    };
  \end{tikzpicture}%
}
\newcommand{\kittireconimgscaledwithinsetbl}[5]{%
  \begin{tikzpicture}[inner sep=0,outer sep=0]
    \node[anchor=south west,inner sep=0] (img) at (0,0) {%
      \includegraphics[width=#3\linewidth,keepaspectratio,trim=#2,clip]{\kittireconpath#1}%
    };
    \node[
      anchor=south west,
      xshift=1.5pt,
      yshift=1.5pt,
      inner sep=0.8pt,
      fill=white,
      draw=black,
      line width=0.25pt
    ] at (img.south west) {%
      \includegraphics[width=#5\linewidth]{#4}%
    };
  \end{tikzpicture}%
}
\newcommand{\kittireconimgscaledwithinsetblpadded}[6]{%
  \begin{tikzpicture}[inner sep=0,outer sep=0]
    \node[anchor=south west,inner sep=0] (frame) at (0,0) {%
      \makebox[#3\linewidth][r]{%
        \includegraphics[width=#4\linewidth,keepaspectratio,trim=#2,clip]{\kittireconpath#1}%
      }%
    };
    \node[
      anchor=north west,
      xshift=1.5pt,
      yshift=-1.5pt,
      inner sep=0.8pt,
      fill=white,
      draw=black,
      line width=0.25pt
    ] at (frame.north west) {%
      \includegraphics[width=#6\linewidth]{#5}%
    };
  \end{tikzpicture}%
}
\newcommand{\kittireconimgscaledwithinsetblpaddedellipse}[9]{%
  \begin{tikzpicture}[inner sep=0,outer sep=0]
    \node[anchor=south west,inner sep=0] (frame) at (0,0) {%
      \makebox[#3\linewidth][r]{%
        \includegraphics[width=#4\linewidth,keepaspectratio,trim=#2,clip]{\kittireconpath#1}%
      }%
    };
    \begin{scope}[shift={(frame.south west)}, x={(frame.south east)}, y={(frame.north west)}]
      \draw[red, dash pattern=on 2pt off 2pt, line cap=round, line width=0.8pt] (#7,#8) ellipse [#9];
    \end{scope}
    \node[
      anchor=north west,
      xshift=1.5pt,
      yshift=-1.5pt,
      inner sep=0.8pt,
      fill=white,
      draw=black,
      line width=0.25pt
    ] at (frame.north west) {%
      \includegraphics[width=#6\linewidth]{#5}%
    };
  \end{tikzpicture}%
}
\newcommand{\kittireconimgscaledwithinsetellipse}[9]{%
  \begin{tikzpicture}[inner sep=0,outer sep=0]
    \node[anchor=south west,inner sep=0] (img) at (0,0) {%
      \includegraphics[width=#3\linewidth,keepaspectratio,trim=#2,clip]{\kittireconpath#1}%
    };
    \begin{scope}[shift={(img.south west)}, x={(img.south east)}, y={(img.north west)}]
      \draw[red, dash pattern=on 2pt off 2pt, line cap=round, line width=0.8pt] (#6,#7) ellipse [x radius=#8, y radius=#9];
    \end{scope}
    \node[
      anchor=north west,
      xshift=1.5pt,
      yshift=-1.5pt,
      inner sep=0.8pt,
      fill=white,
      draw=black,
      line width=0.25pt
    ] at (img.north west) {%
      \includegraphics[width=#5\linewidth]{#4}%
    };
  \end{tikzpicture}%
}
\newcommand{\kittireconimgscaledwithinsetellipsepadded}[9]{%
  \begin{tikzpicture}[inner sep=0,outer sep=0]
    \node[
      anchor=south west,
      inner sep=0,
      outer sep=0,
      minimum width=#3\linewidth
    ] (frame) at (0,0) {%
      \includegraphics[width=#4\linewidth,keepaspectratio,trim=#2,clip]{\kittireconpath#1}%
    };
    \begin{scope}[shift={(frame.south west)}, x={(frame.south east)}, y={(frame.north west)}]
      \draw[red, dash pattern=on 2pt off 2pt, line cap=round, line width=0.8pt] (#7,#8) ellipse [#9];
    \end{scope}
    \node[
      anchor=north west,
      xshift=1.5pt,
      yshift=-1.5pt,
      inner sep=0.8pt,
      fill=white,
      draw=black,
      line width=0.25pt
    ] at (frame.north west) {%
      \includegraphics[width=#6\linewidth]{#5}%
    };
  \end{tikzpicture}%
}
\newcommand{\kittireconimgwithinsettl}[4]{%
  \begin{tikzpicture}[inner sep=0,outer sep=0]
    \node[anchor=south west,inner sep=0] (img) at (0,0) {%
      \includegraphics[width=\linewidth,keepaspectratio,trim=#2,clip]{\kittireconpath#1}%
    };
    \node[
      anchor=north west,
      xshift=1.5pt,
      yshift=-1.5pt,
      inner sep=0.8pt,
      fill=white,
      draw=black,
      line width=0.25pt
    ] at (img.north west) {%
      \includegraphics[width=#4\linewidth]{#3}%
    };
  \end{tikzpicture}%
}
\newcommand{\kittireconimgpaddedtopwithinsetbc}[6]{%
  \begin{tikzpicture}[inner sep=0,outer sep=0]
    \node[
      anchor=south west,
      inner sep=0,
      outer sep=0,
      minimum width=\linewidth,
      minimum height=#3,
      fill=white
    ] (frame) at (0,0) {};
    \node[anchor=south west,inner sep=0,outer sep=0] at (frame.south west) {%
      \makebox[\linewidth][c]{%
        \includegraphics[width=#4\linewidth,keepaspectratio,trim=#2,clip]{\kittireconpath#1}%
      }%
    };
    \node[
      anchor=south,
      xshift=0pt,
      yshift=1.5pt,
      inner sep=0.8pt,
      fill=white,
      draw=black,
      line width=0.25pt
    ] at (frame.south) {%
      \includegraphics[width=#6\linewidth]{#5}%
    };
  \end{tikzpicture}%
}
\newcommand{\kittireconimgpaddedtopwithinsettl}[6]{%
  \begin{tikzpicture}[inner sep=0,outer sep=0]
    \node[
      anchor=south west,
      inner sep=0,
      outer sep=0,
      minimum width=\linewidth,
      minimum height=#3,
      fill=white
    ] (frame) at (0,0) {};
    \node[anchor=south west,inner sep=0,outer sep=0] at (frame.south west) {%
      \makebox[\linewidth][c]{%
        \includegraphics[width=#4\linewidth,keepaspectratio,trim=#2,clip]{\kittireconpath#1}%
      }%
    };
    \node[
      anchor=north west,
      xshift=1.5pt,
      yshift=-1.5pt,
      inner sep=0.8pt,
      fill=white,
      draw=black,
      line width=0.25pt
    ] at (frame.north west) {%
      \includegraphics[width=#6\linewidth]{#5}%
    };
  \end{tikzpicture}%
}
\newcommand{\kittireconimgpaddedtopwithinsettlellipse}[9]{%
  \begin{tikzpicture}[inner sep=0,outer sep=0]
    \node[
      anchor=south west,
      inner sep=0,
      outer sep=0,
      minimum width=\linewidth,
      minimum height=#3,
      fill=white
    ] (frame) at (0,0) {};
    \node[anchor=south,inner sep=0,outer sep=0] (img) at (frame.south) {%
      \includegraphics[width=#4\linewidth,keepaspectratio,trim=#2,clip]{\kittireconpath#1}%
    };
    \begin{scope}[shift={(img.south west)}, x={(img.south east)}, y={(img.north west)}]
      \draw[red, dash pattern=on 2pt off 2pt, line cap=round, line width=0.8pt] (#7,#8) ellipse [#9];
    \end{scope}
    \node[
      anchor=north west,
      xshift=1.5pt,
      yshift=-1.5pt,
      inner sep=0.8pt,
      fill=white,
      draw=black,
      line width=0.25pt
    ] at (frame.north west) {%
      \includegraphics[width=#6\linewidth]{#5}%
    };
  \end{tikzpicture}%
}
\newcommand{\kittireconimgwithinsettlellipse}[8]{%
  \begin{tikzpicture}[inner sep=0,outer sep=0]
    \node[anchor=south west,inner sep=0] (img) at (0,0) {%
      \includegraphics[width=\linewidth,keepaspectratio,trim=#2,clip]{\kittireconpath#1}%
    };
    \begin{scope}[shift={(img.south west)}, x={(img.south east)}, y={(img.north west)}]
      \draw[red, dash pattern=on 2pt off 2pt, line cap=round, line width=0.8pt] (#5,#6) ellipse [x radius=#7, y radius=#8];
    \end{scope}
    \node[
      anchor=north west,
      xshift=1.5pt,
      yshift=-1.5pt,
      inner sep=0.8pt,
      fill=white,
      draw=black,
      line width=0.25pt
    ] at (img.north west) {%
      \includegraphics[width=#4\linewidth]{#3}%
    };
  \end{tikzpicture}%
}
\newcommand{\kittireconplaceholder}[1]{%
  \fbox{\parbox[c][0.19\textwidth][c]{\dimexpr\linewidth-2\fboxsep-2\fboxrule\relax}{\centering \footnotesize #1\\(placeholder)}}%
}

\begin{strip}
\vspace{-4.5em}
  \centering

  \begin{minipage}[t]{0.33\textwidth}
    \centering
    \kittireconimgscaledwithinsetblpaddedellipse{ours_seq00_cam2}{0 0 0 0}{1.0}{0.67}{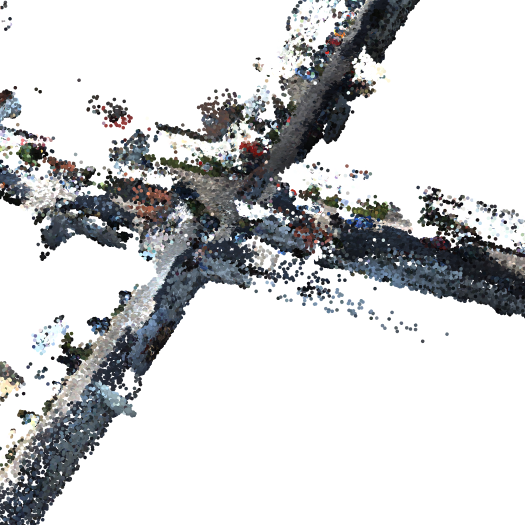}{0.23}{0.60}{0.75}{x radius=0.030, y radius=0.081}
    \\[-0.4ex]\footnotesize KITTI-00 (3724m)
  \end{minipage}\hfill
  \begin{minipage}[t]{0.33\textwidth}
    \centering
    \kittireconimgscaledwithinsetellipsepadded{ours_seq01_cam2}{0 0 0 0}{1.0}{0.69}{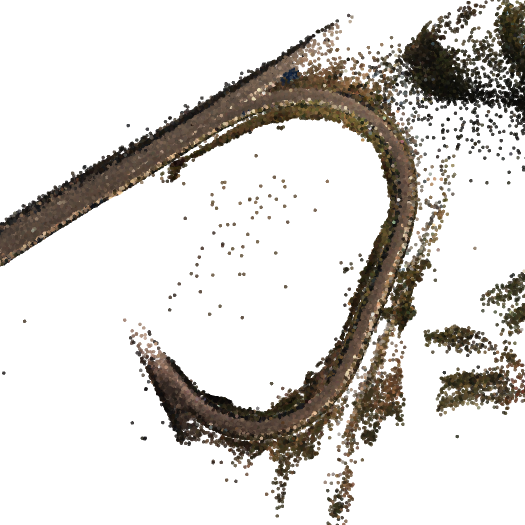}{0.23}{0.757}{0.783}{x radius=0.037, y radius=0.093}
    \\[-0.4ex]\footnotesize KITTI-01 (2453m)
  \end{minipage}\hfill
  \begin{minipage}[t]{0.33\textwidth}
    \centering
    \kittireconimgwithinsettlellipse{ours_seq02_cam2}{0 0 0 0}{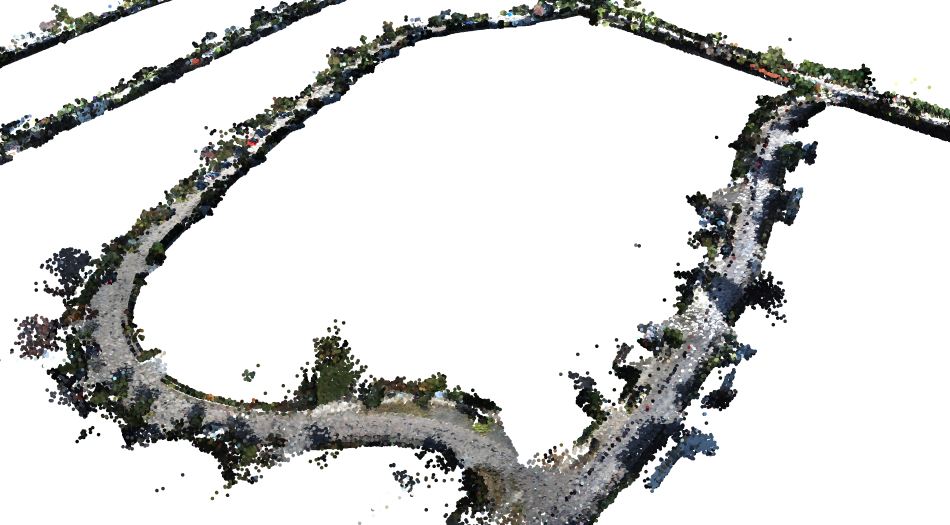}{0.34}{0.7}{0.75}{0.12}{0.20}
    \\[-0.4ex]\footnotesize KITTI-02 (5067m)
  \end{minipage}

  \vspace{0.2cm}

  \begin{minipage}[t]{0.33\textwidth}
    \centering
    \kittireconimgpaddedtopwithinsettrellipse{ours_seq05_cam2}{0bp 0bp 0bp 0bp}{\kittireconrowtwoheight}{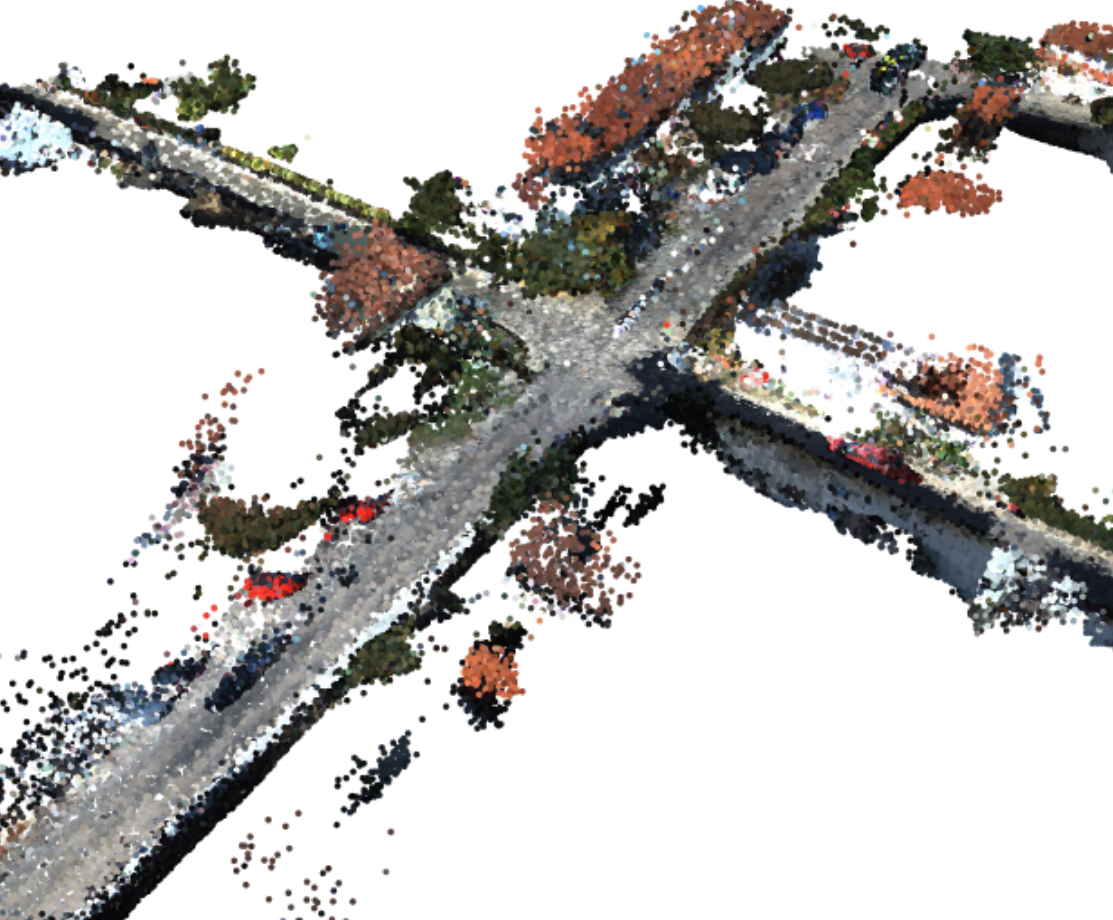}{0.23}{0.50}{0.72}{x radius=0.08, y radius=0.1352}
    \\[-0.4ex]\footnotesize KITTI-05 (2205m)
  \end{minipage}\hfill
  \begin{minipage}[t]{0.33\textwidth}
    \centering
    \kittireconimgpaddedtopwithinsettlellipse{ours_seq07_cam2}{0 0 0 0}{\kittireconrowtwoheight}{0.93}{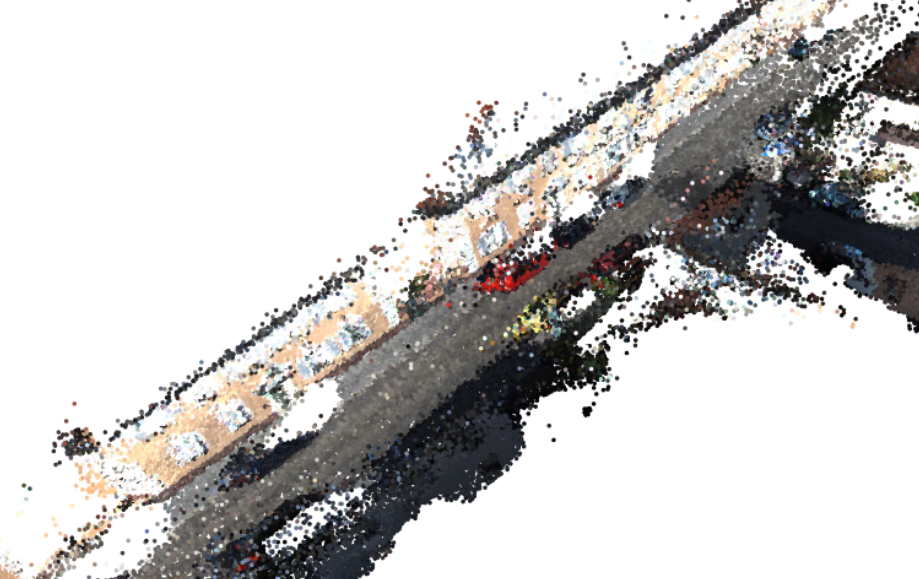}{0.23}{0.394}{0.497}{x radius=0.086, y radius=0.124}
    \\[-0.4ex]\footnotesize KITTI-07 (649m)
  \end{minipage}\hfill
  \begin{minipage}[t]{0.33\textwidth}
    \centering
    \kittireconimgpaddedtopwithinsettrellipse{ours_seq08_cam2}{0 0 0 0}{\kittireconrowtwoheight}{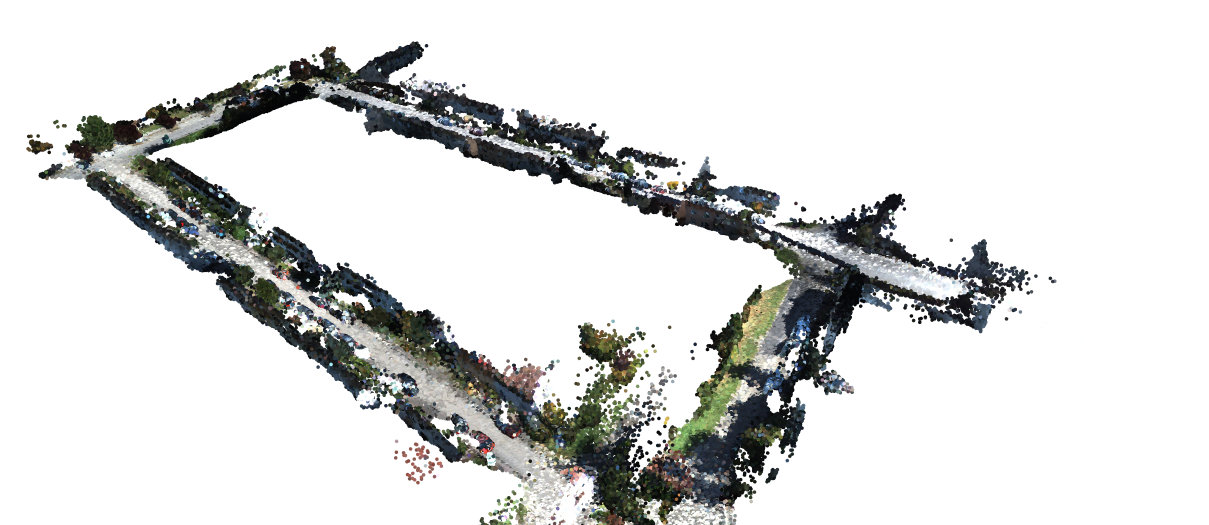}{0.4}{0.2}{0.85}{x radius=0.085, y radius=0.125}
    \\[-0.4ex]\footnotesize KITTI-08 (3222m)
  \end{minipage}

  \captionof{figure}{The kilometer-scale \textbf{\ours-Long} reconstructions use our feedforward predictions with segments of 120 frames, followed by the trajectory-level post-optimization of \cite{deng2025vggtlong}; red circles show enlarged local point maps.}
  \label{fig:teaser-kitti-recon}
\end{strip}

\begin{abstract}
Stable and reliable 4D spatial understanding is fundamental for autonomous driving systems.
While feedforward reconstruction networks can estimate camera motion and 3D structure in one pass,
pose estimation over long videos remains challenged by computational cost, long-context ambiguity, and temporal instability.
To address these challenges, we propose Feedforward Visual Odometry (FFVO), a pose-specialized adaptation of joint reconstruction architectures for efficient and temporally stable camera-pose estimation.
FFVO uses (i) a compact camera-token representation for computationally efficient temporal aggregation, (ii) a hierarchical local-to-global temporal decoder that mitigates geometric ambiguity by separating short-range motion aggregation from sequence-level integration, and (iii) intermediate trajectory supervision that promotes temporal stability.
Extensive evaluation on the Waymo Open Dataset (WOD), KITTI, and a large-scale proprietary benchmark demonstrates that our method performs favorably against existing feedforward approaches, and greatly reduces jitter and drift. 
These results support FFVO as an effective feedforward camera-pose decoder in long-horizon visual odometry settings.
\end{abstract}

\section{INTRODUCTION}
\begin{figure}[t]
  \centering
  \vspace{0.5em} %
  \begin{subfigure}[t]{0.245\linewidth}
    \centering
    \includegraphics[width=\linewidth]{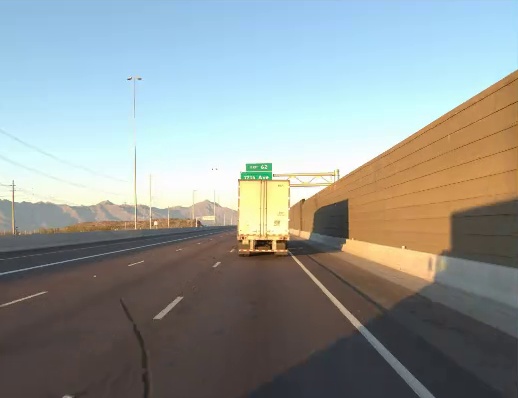}\\[0.4ex]
    \includegraphics[width=\linewidth]{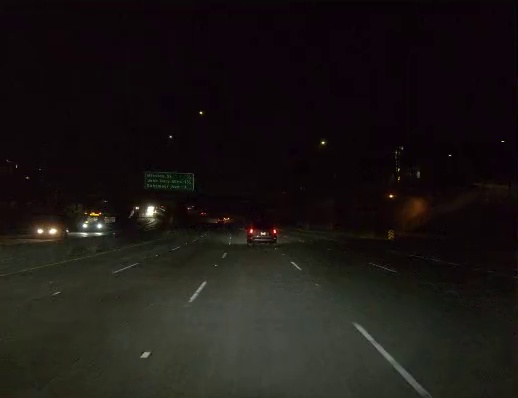}
    \captionsetup{font=footnotesize}
    \caption{Input video}
    \label{fig:motivation_input}
  \end{subfigure}\hfill
  \begin{subfigure}[t]{0.245\linewidth}
    \centering
    \includegraphics[width=\linewidth]{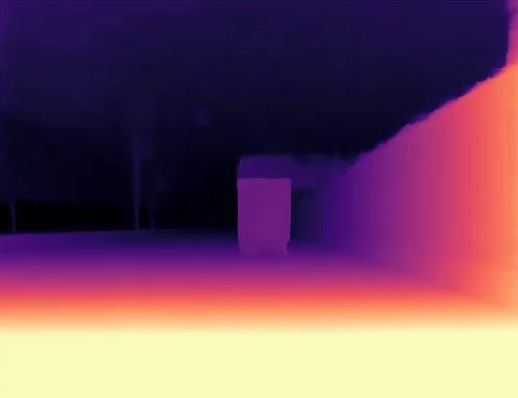}\\[0.4ex]
    \includegraphics[width=\linewidth]{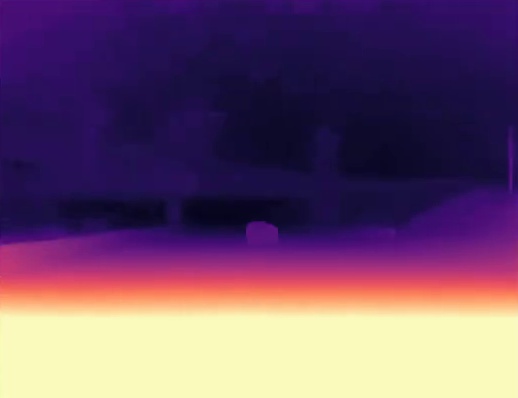}
    \captionsetup{font=footnotesize, justification=centering}
    \caption{Point map ($z$-axis)}
    \label{fig:motivation_depth}
  \end{subfigure}\hfill
  \begin{subfigure}[t]{0.245\linewidth}
    \centering
    \includegraphics[width=\linewidth]{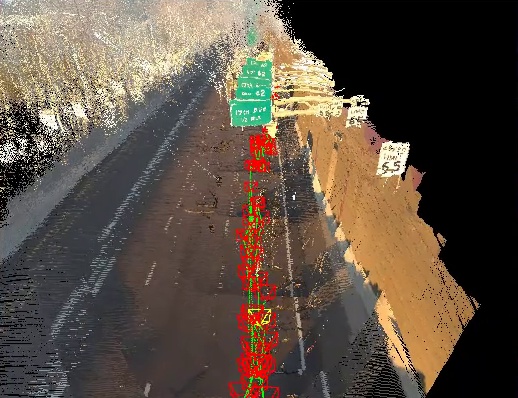}\\[0.4ex]
    \includegraphics[width=\linewidth]{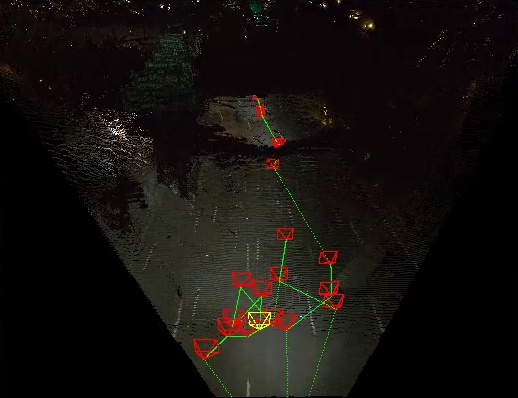}
    \captionsetup{font=footnotesize}
    \caption{\pithree~\cite{wang2025pi3}}
    \label{fig:motivation_pi3}
  \end{subfigure}\hfill
  \begin{subfigure}[t]{0.245\linewidth}
    \centering
    \includegraphics[width=\linewidth]{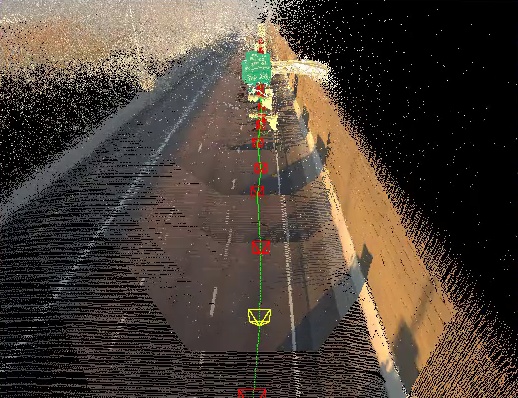}\\[0.4ex]
    \includegraphics[width=\linewidth]{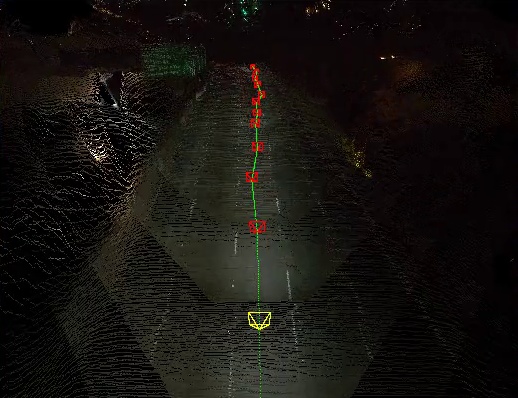}
    \captionsetup{font=footnotesize}
    \caption{\ourspost}
    \label{fig:motivation_ours}
  \end{subfigure}
  \caption{
\textbf{Feedforward reconstruction.}
As shown in (b), the state-of-the-art feedforward model \pithree~\cite{wang2025pi3} is able to produce locally accurate geometry.
However, it exhibits pose jitter that misaligns geometry across frames, leading to duplicated static structures (e.g., road signs and barriers) when given long-horizon videos as input, as shown in (c).
This observation motivates us to develop a better pose regression branch to improve long term temporal consistency.
As shown in (d), our reconstruction is more temporally coherent. Camera poses: yellow denotes the current frame, red denotes other frames. We remove dynamic agents and sky for better visualization.
}
  
  \label{fig:motivation_placeholder}
\end{figure}

Accurate and stable camera pose estimation is a foundational capability for
autonomous driving, enabling mapping, localization, motion forecasting,
long-horizon scene understanding \cite{geiger2012kitti,sun2020waymo}, and transforms uncurated videos (e.g., internet footage) into paired trajectory-video data for training large-scale world models for driving~\cite{jiang_waymo_2026,ren2025cosmos}.
While classical and learning-based Visual Odometry (VO) and SLAM pipelines have
achieved strong performance
\cite{czarnowski2020deepfactors,teed2021droidslam,teed2023deep}, modern
self-driving systems increasingly benefit from 3D foundation models
that can exploit large-scale data and jointly reason about geometry, appearance,
and dynamics.
Recently, transformer-based models have shown that it is possible to predict
camera motion and 3D structure simultaneously \cite{wang2025vggt,wang2025pi3,lin2025da3,keetha2026mapanything}.
By jointly regressing camera poses and dense geometric fields, such as depth or point maps, in a single network feedforward pass, these architectures offer a scalable alternative to iterative multi-view optimization~\cite{teed2021droidslam,mur2017orb,gao2018ldso}. However, while these feedforward predictions are locally accurate, constructing a globally coherent map still hinges on precise temporal alignment across the entire trajectory. In long-horizon self-driving scenarios, these camera pose estimates often become unstable, exhibiting jitter, fragmentation, or drift. As shown in \figref{fig:motivation_placeholder}, the resulting misalignment breaks cross-frame coherency of the reconstructed point map.

This gap between local geometric quality and global trajectory stability is particularly prevalent in self-driving settings, where trajectories often span hundreds to thousands of frames.
However, directly extending transformer-based models to these long horizons is often prohibitive; the computational complexity of standard attention mechanisms~\cite{vaswani2017attention} scales quadratically with sequence length, and unconstrained global attention can become ambiguous due to repeated scene structures—such as lane markings or vegetation—as well as motion blur and dynamic agents. These factors motivate the need for a pose decoding mechanism that leverages long-range context while remaining computationally efficient and locally stable.

To bridge this gap, we introduce \textbf{F}eed\textbf{F}orward \textbf{V}isual \textbf{O}dometry (\textbf{\ours}), a pose-specialized adaptation of joint reconstruction models featuring a feedforward camera-pose decoder for long-horizon visual odometry in autonomous driving. Starting from a pretrained feedforward architecture (e.g., \pithree-style models), we freeze the geometric feature aggregation module and train a dedicated camera pose branch from scratch. This design preserves the model’s learned local geometric representation while re-structuring the pose head to handle the temporal complexities of long trajectories.

Our approach, \textbf{\ours}, centers on three architectural innovations within the pose branch:
First, we represent each frame with a small set of camera tokens that act as compact pose queries, enabling the model to scale to long sequences by avoiding dense all-to-all attention.
Second, we adopt a local-to-global temporal decoder where tokens first aggregate information within a short temporal window to capture stable motion cues, improving trajectory stability and reducing geometric ambiguity.

Third, we enforce temporal stability through intermediate local-trajectory supervision, applying auxiliary relative-pose losses to encourage locally consistent motion estimates.
Together, these modifications produce a pose decoder that is efficient and robust to the drift and jitters typical of driving sequences, and improve ultra long range trajectory fidelity when paired with a post-optimization technique~\cite{deng2025vggtlong}  (\figref{fig:teaser-kitti-recon}). %

Our main contributions are summarized as follows:
\begin{list}{\textbullet}{%
  \setlength{\leftmargin}{1.2em}%
  \setlength{\labelwidth}{0.7em}%
  \setlength{\labelsep}{0.5em}%
  \setlength{\itemsep}{0pt}%
  \setlength{\parsep}{0pt}%
  \setlength{\topsep}{0pt}%
  \setlength{\partopsep}{0pt}%
}
\item \textbf{Efficient camera token design to reduce complexity:} We introduce a compact set of camera tokens as an alternative to dense temporal attention, enabling the model to efficiently process longer trajectories with lower computational complexity.
\item \textbf{Local-to-global temporal attention to address ambiguity:} We propose a hierarchical strategy that combines local-window attention, improving long-range consistency while mitigating geometric ambiguity in long-horizon sequences and reducing the jitters and drift.
\item \textbf{Intermediate local trajectory supervision to improve stability:} We employ auxiliary relative-pose losses at intermediate layers to alleviate temporal instability.
\end{list}

\begin{figure*}[t]
\centering
\includegraphics[width=0.7725\textwidth]{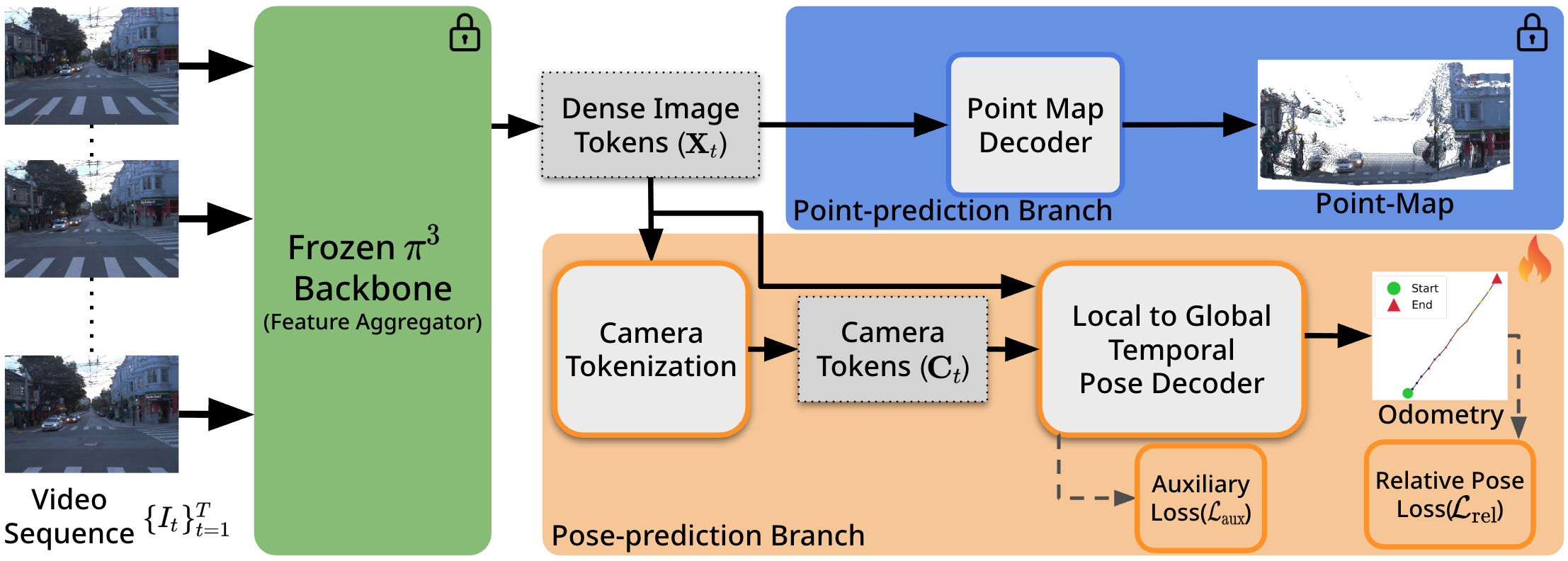}
\caption{
\textbf{Overview.}
A frozen $\pi^3$~\cite{wang2025pi3} backbone produces image tokens and geometry, while the trainable camera-token and local-to-global pose branches predict long-horizon odometry using relative-pose and intermediate local-trajectory supervision.
}
\label{fig:method-overview}
\end{figure*}

\section{Related Work}
\subsection{Joint pose and geometry foundation models}

Recent feedforward reconstruction models unify camera pose estimation with dense geometry
prediction~\cite{wang2026vggt}. VGGT~\cite{wang2025vggt} predicts camera parameters together with point maps, depth, and tracks from
one or many views in a single transformer forward pass. Following this, $\pi^3$~\cite{wang2025pi3} adopts a permutation-equivariant formulation to predict poses and scale-invariant local point maps
without selecting a reference view. Depth Anything 3~\cite{lin2025da3} extends the any-view
setting by predicting spatially consistent geometry from an arbitrary number of inputs, with
or without known poses \cite{lin2025da3}. For autonomous driving, several works adapt these
geometry-token paradigms to multi-camera vehicle setups and driving priors, including DriveVGGT~\cite{jia2025drivevggt} and DVGT~\cite{zuo2025dvgt}. Beyond static scenes, PAGE-4D~\cite{zhou2025page4d} and D4RT~\cite{zhang2026d4rt}
address pose and geometry estimation in dynamic videos.

Parallel research addresses the scalability of geometric transformers for long sequences through both architectural efficiency and system-level integration~\cite{cheng2026horizonstream,tao2026anchor3r,xu2026framevggt,deng2026mamba,sun2026nodrift3r}. On the model level, Faster-VGGT \cite{wang2025fastervggt} and Co-Me \cite{chen2025come} mitigate the quadratic cost of attention through block-sparse kernels and confidence-guided token merging, whereas LASER \cite{ding2025laser} introduces layer-wise scale alignment for streaming stability. Conversely, frameworks such as VGGT-Long \cite{deng2025vggtlong} and VGGT-SLAM 2.0 \cite{maggio2026vggtslam2} introduce post-hoc system-level wrappers—including submap alignment and $SL(4)$ manifold optimization—to achieve kilometer-scale reconstruction~\cite{zhang2026vggt}. Our \ours{} embeds temporal consistency within the transformer architecture to improve feedforward pose estimates within each input segment. For kilometer-scale trajectories, \ours-Long combines these segment-level predictions using the post-optimization pipeline of VGGT-Long~\cite{deng2025vggtlong}.

\subsection{Learned visual odometry and SLAM for long horizons}

Learned long-horizon VO/SLAM methods commonly combine neural matching with geometric optimization. 
DROID-SLAM~\cite{teed2021droidslam} performs recurrent correspondence refinement with dense 
differentiable bundle adjustment (BA) for global consistency, while DPVO~\cite{teed2023deep} 
improves efficiency via sparse patch tracking plus differentiable BA. Building on DPVO, 
DPV-SLAM and DPV-SLAM++~\cite{lipson2024deep} add SLAM-level optimization to improve 
long-sequence robustness, and MASt3R-SLAM~\cite{murai2024mast3rslam} leverages 
MASt3R~\cite{leroy2024grounding} learned 3D reconstruction priors for dense SLAM. 
In contrast, \ours\ focuses on building long-range temporal consistency directly into the 
network through hierarchical pose decoding with camera tokens and intermediate supervision.

\subsection{Transfomer-based efficient long-sequence pose decoding}
Transformers have increasingly been applied to pose estimation and visual odometry 
through attention-based motion aggregation~\cite{lai2025zerovo, dong2024reloc3r, kurt2024causalvio, wang2025vggt,wang2025pi3}. 
However, as trajectories grow, standard global attention incurs rapidly increasing 
computational cost~\cite{wang2025vggt,wang2025pi3} and can become ambiguous 
in long-horizon driving scenes, especially under repeated structures, motion blur, 
and dynamic agents.

To maintain scalability, long-sequence modeling often combines local-window attention 
with global interactions, as introduced in Longformer \cite{beltagy2020longformer} 
and Video Swin Transformer \cite{liu2021videoswin}. While originally developed for text 
or dense spatiotemporal patches, these ideas transfer naturally to long-horizon pose decoding. 
Our \ours{} adapts this idea to visual odometry by using compact camera tokens and a hierarchical local-to-global temporal decoder to 
first aggregate stable short-range motion cues, then integrate trajectory-level context 
to reduce long-horizon ambiguity and improve temporal consistency. This design avoids dense 
all-to-all temporal attention while preserving efficiency for long-trajectory odometry.

\section{Method}
\label{sec:method}

\subsection{Overview}
\figref{fig:method-overview} shows an overview of \ours. Given a monocular video sequence $\{I_t\}_{t=1}^{T}$, \ours{} predicts a camera-to-world pose
$\mathbf{T}_{w\leftarrow c_t}\in SE(3)$ for each frame in a feedforward manner, without
assuming known camera intrinsics. We adopt a pretrained \pithree\ model \cite{wang2025pi3},
which produces per-frame visual tokens and geometric predictions in an uncalibrated, reference-free
setting. We take and freeze \pithree's feature aggregation and point-map pathway as our backbone, and train our pose decoding branch, designed for long-horizon driving trajectories, from scratch. The pose branch uses (i) a small
set of \emph{camera tokens} per frame as compact pose queries, (ii) a hierarchical local-window temporal cross-attention
followed by a global integration stage, and (iii) a relative-pose training objective with intermediate
local-trajectory supervision.

\begin{figure}[t]
\centering
\includegraphics[width=0.925\linewidth]{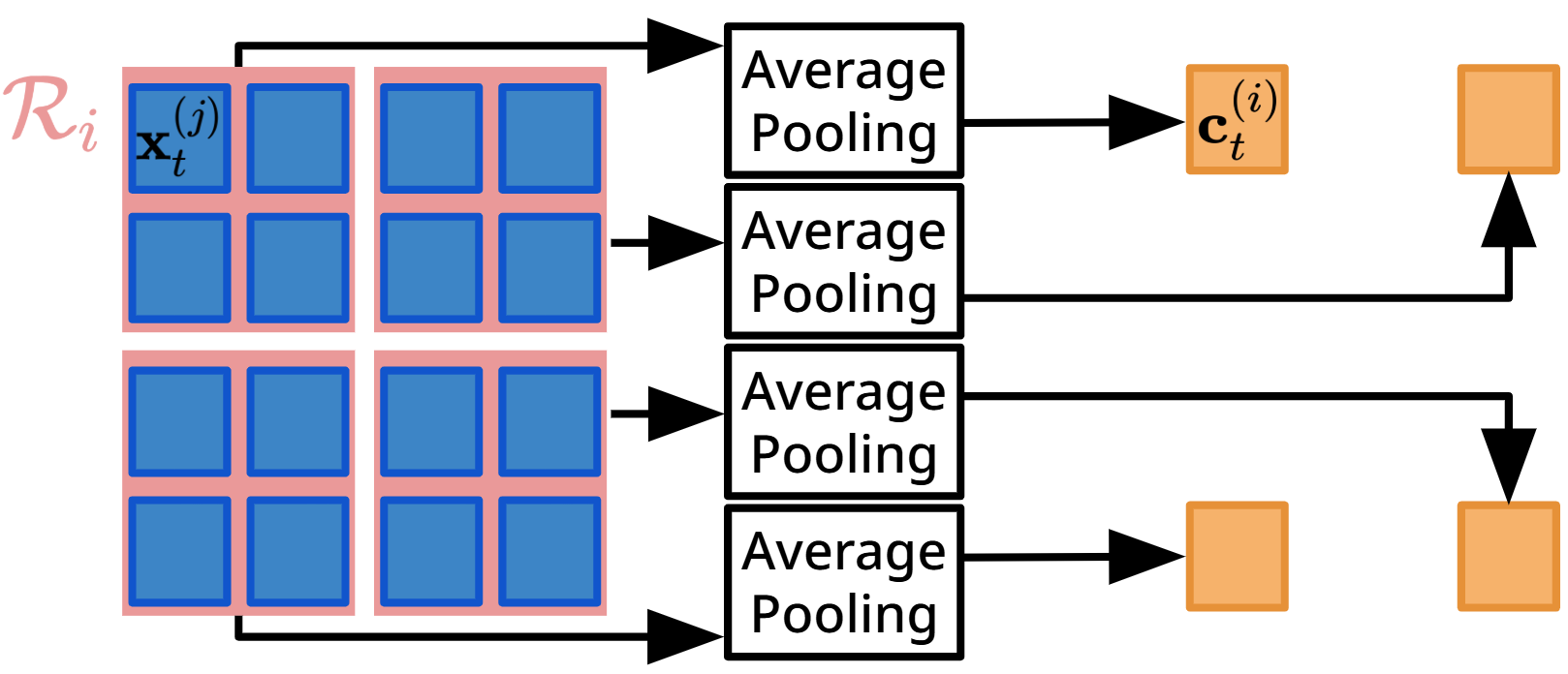}
\caption{
\textbf{Camera tokenization.}
For each frame $t$, we partition the frozen image tokens $\mathbf{X}_t=\{\mathbf{x}_t^{(j)}\}_{j=1}^{N}$ into $M$ coarse regions $\mathcal{R}_i$ and average-pool $\mathbf{x}_t^{(j)}$ within each region to initialize camera tokens $\mathbf{C}_t=\{\mathbf{c}_t^{(i)}\}_{i=1}^{M}$, which serve as compact pose queries for temporal
cross-attention.
}
\label{fig:camera-tokens}
\end{figure}

\subsection{Backbone and frozen components}
Let $\mathbf{X}_t=\{\mathbf{x}_t^{(j)}\}_{j=1}^{N}\in\mathbb{R}^{N\times D}$ denote the per-frame image
tokens produced by the frozen \pithree\ backbone for frame $I_t$, where
$\mathbf{x}_t^{(j)}\in\mathbb{R}^{D}$ (with $N$ tokens of dimension $D$). We keep the feature aggregation
stack and the point-map (geometry) pathway fixed, and update only the parameters of the pose decoding
branch described in \secref{sec:local-to-global-decoder}.

\subsection{Camera tokenization}
\label{sec:camera-tokenization}
For each frame $t$, we introduce $M$ camera tokens $\mathbf{C}_t=\{\mathbf{c}_t^{(i)}\}_{i=1}^{M}$ with
$\mathbf{c}_t^{(i)}\in\mathbb{R}^{D}$. We initialize camera tokens by partitioning the image into a coarse
grid of $M$ regions and average-pooling the frozen image tokens within each region, yielding a compact set ($M\!\ll\!N$). Writing
$\mathcal{R}_i$ for the set of image-token indices in region $i$, the initialization is (\figref{fig:camera-tokens})
\begin{equation}
\mathbf{c}_t^{(i)} \;=\; \frac{1}{|\mathcal{R}_i|}\sum_{j\in\mathcal{R}_i}\mathbf{x}_t^{(j)},\qquad i=1,\dots,M.
\end{equation}
These tokens act as a compact representation of the camera state that also captures region-specific features, and are used as queries for temporal aggregation.

\subsection{Local-to-global temporal pose decoder}
\label{sec:local-to-global-decoder}
\begin{figure}[t]%
    \centering%
    \begin{subfigure}[c]{0.365\linewidth}%
        \centering%
        \vspace{0.85em} %
        \parbox[t]{\linewidth}{%
            \includegraphics[width=\linewidth]{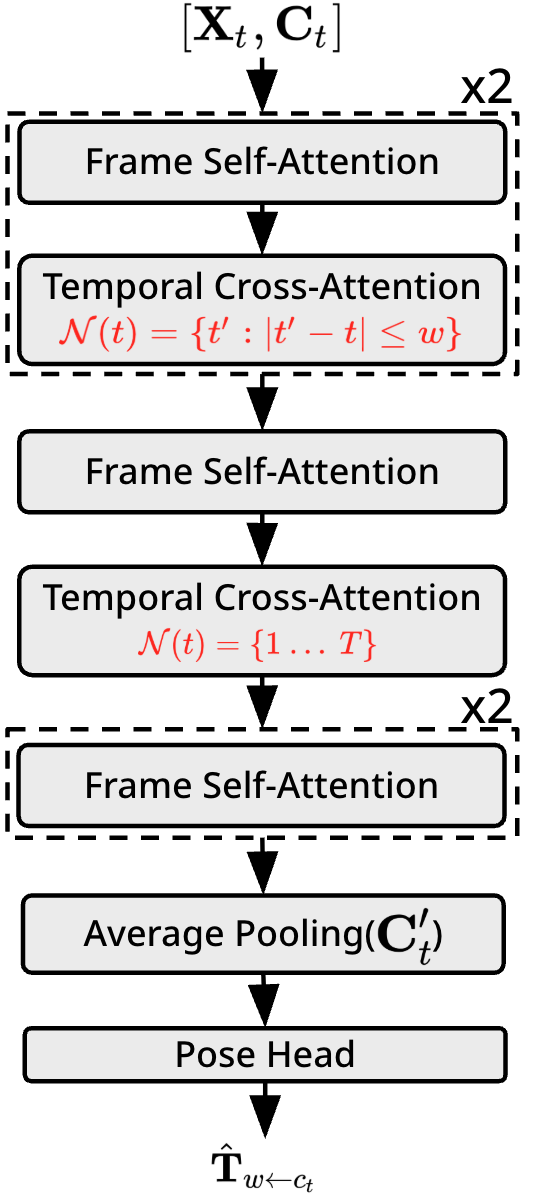}%
        }%
        \caption{Local-to-global decoder architecture}%
        \label{fig:arch}%
    \end{subfigure}%
    \hfill%
    \begin{subfigure}[c]{0.60\linewidth}
        \centering
        \begin{subfigure}{\linewidth}%
            \centering%
            \parbox[t]{\linewidth}{%
                \includegraphics[width=\linewidth]{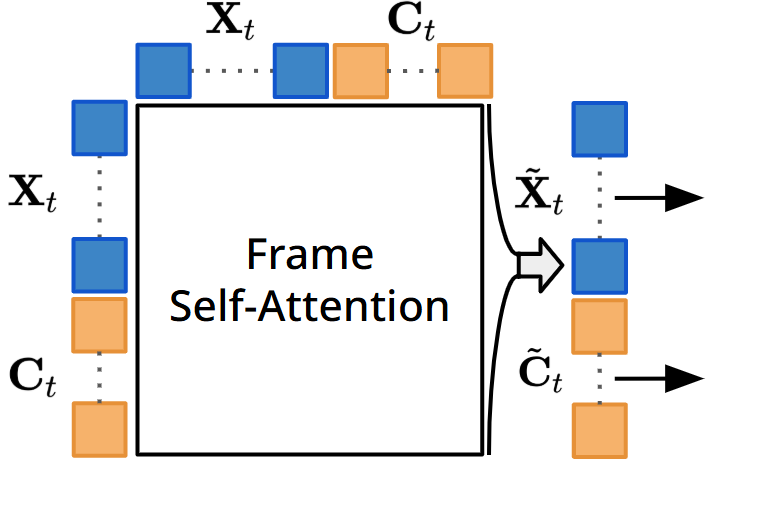}
            }%
            \caption{Frame self-attention block}%
            \label{fig:self-attn}%
        \end{subfigure}
        \begin{subfigure}{\linewidth}%
            \centering%
            \includegraphics[width=\linewidth]{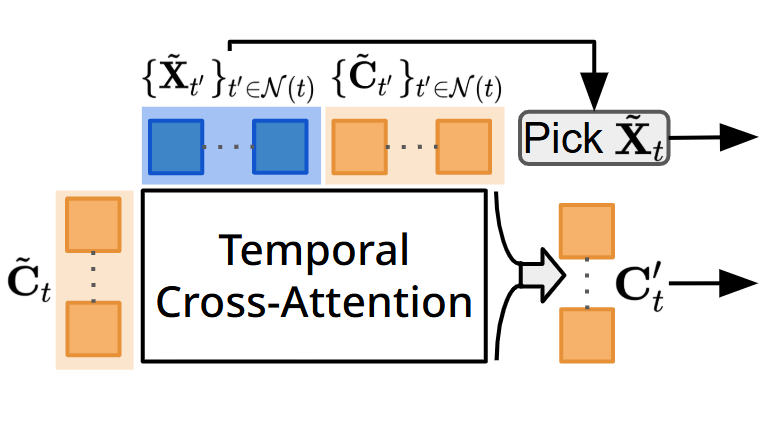}
            \caption{Temporal cross-attention block}%
            \label{fig:cross-attn}%
        \end{subfigure}%
    \end{subfigure}%
    \caption{
    \textbf{Local-to-global temporal pose decoder} with its overall architecture and key attention modules. (a) Overall architecture with two local-window stages and one global integration stage. (b) Frame self-attention within a single frame. (c) Temporal cross-attention where camera tokens query temporal context.
    }%
    \label{fig:local-to-global}%
\end{figure}

We update both image and camera tokens by interleaving (i) \emph{frame self-attention} within each frame and (ii) \emph{temporal
cross-attention} across neighboring frames (\figref{fig:arch}). Let $\tilde{\mathbf{X}}_t,\tilde{\mathbf{C}}_t$ be the updated
tokens after one frame-level block (\figref{fig:self-attn}):
\begin{equation}
[\tilde{\mathbf{X}}_t;\tilde{\mathbf{C}}_t] \;=\; \mathrm{Attn}_{\mathrm{frame}}\big([\mathbf{X}_t;\mathbf{C}_t]\big),
\end{equation}
where attention is restricted to tokens from the same frame. We then perform temporal cross-attention in
which only camera tokens query a temporal neighborhood $\mathcal{N}(t)=\{t':|t'-t|\leq w\}$ (\figref{fig:cross-attn}):
\begin{equation}
\begin{aligned}
\mathbf{Y}_t &= \mathrm{Concat}\!\left(\{[\tilde{\mathbf{X}}_{t'};\tilde{\mathbf{C}}_{t'}]\}_{t'\in\mathcal{N}(t)}\right),\\
\mathbf{C}'_t &= \mathrm{Attn}_{\mathrm{temp}}\!\left(\mathbf{Q}=\tilde{\mathbf{C}}_t,\mathbf{K}=\mathbf{Y}_t,\mathbf{V}=\mathbf{Y}_t\right).
\end{aligned}
\end{equation}
This design avoids dense token-to-token temporal attention: only the $M$ camera tokens (a fixed
hyperparameter) are used as temporal queries, and each query attends to at most
$T(N\!+\!M)$ keys/values in the final global stage. Therefore the
temporal cross-attention compute is
$\mathcal{O}\!\left(T^2\,M\,(N\!+\!M)\right)$; with fixed $M$ and $N\!\gg\!M$, this simplifies to
$\mathcal{O}\!\left(T^2N\right)$ computationally, compared with dense all-image-token temporal self-attention
$\mathcal{O}\!\left(T^2N^2\right)$.

\subsection{Pose parameterization and readout}
For each frame, we pool camera tokens into a single representation $\bar{\mathbf{c}}_t=\frac{1}{M}\sum_{i=1}^M
\mathbf{c}'^{(i)}_t$ and apply a lightweight MLP head identical to \cite{wang2025pi3} to predict translation $\hat{\mathbf{t}}_t\in\mathbb{R}^3$
and a 6D rotation representation $\hat{\mathbf{r}}_t\in\mathbb{R}^6$ \cite{zhou2019continuity}. The 6D
representation is mapped to $\hat{\mathbf{R}}_t\in SO(3)$ via normalization and orthogonalization, and we assemble
\begin{equation}
\hat{\mathbf{T}}_{w\leftarrow c_t} \;=\;
\begin{bmatrix}
\hat{\mathbf{R}}_t & \hat{\mathbf{t}}_t\\
\mathbf{0}^\top & 1
\end{bmatrix}.
\end{equation}

\subsection{Training objectives}
\textbf{Relative pose supervision.} Following \pithree\ \cite{wang2025pi3}, we supervise relative motion to avoid
fixing any frame as the reference. Given predicted absolute poses, the relative transform from $t$ to $t'$ is
\begin{equation}
\hat{\mathbf{T}}_{c_{t'}\leftarrow c_t} \;=\; \hat{\mathbf{T}}_{w\leftarrow c_{t'}}^{-1}\hat{\mathbf{T}}_{w\leftarrow c_t},
\end{equation}
and we compare it against the ground-truth relative transform $\mathbf{T}_{c_{t'}\leftarrow c_t}$ for pairs
$(t,t')\in\mathcal{P}$. For brevity, we write $\hat{\mathbf{T}}_{t'\leftarrow t}\triangleq\hat{\mathbf{T}}_{c_{t'}\leftarrow c_t}$
and $\mathbf{T}_{t'\leftarrow t}\triangleq\mathbf{T}_{c_{t'}\leftarrow c_t}$. We use a geodesic rotation distance
$d_R(\hat{\mathbf{R}},\mathbf{R})=\arccos\!\left(\frac{\mathrm{tr}(\hat{\mathbf{R}}^\top \mathbf{R})-1}{2}\right)$ (the angular distance on $SO(3)$)
and an $\ell_1$ translation loss:
\begin{equation}
\begin{aligned}
\mathcal{L}_{\mathrm{rel}} \;=\; \sum_{(t,t')\in\mathcal{P}} \Big(
&\lambda_R\, d_R\!\left(\hat{\mathbf{R}}_{t'\leftarrow t},\mathbf{R}_{t'\leftarrow t}\right)\\
&+\lambda_t\, \left\|\hat{\mathbf{t}}_{t'\leftarrow t}-\mathbf{t}_{t'\leftarrow t}\right\|_1
\Big).
\end{aligned}
\label{eq:rel-loss}
\end{equation}
where $\lambda_R$ and $\lambda_t$ are weighting hyperparameters.

\textbf{Intermediate local-trajectory supervision.} To improve temporal stability, we attach auxiliary pose heads
after the second local-window temporal cross-attention stage. Specifically, we average the camera tokens within
each frame to obtain $\bar{\mathbf{c}}_t^{(2)}$ and use a linear layer to predict a \emph{local trajectory} of
relative poses over the same temporal neighborhood $\mathcal{N}(t)$. This auxiliary head encourages each frame's
camera tokens to encode both its own state and the motion of nearby frames. We apply the same relative-pose
supervision over the predicted local trajectory:
\begin{equation}
\begin{aligned}
\mathcal{L}_{\mathrm{aux}} &= \sum_{t=1}^{T}\sum_{t'\in\mathcal{N}(t)}
\Big(
\lambda_R\, d_R\!\left(\hat{\mathbf{R}}^{\mathrm{aux}}_{t'\leftarrow t},\mathbf{R}_{t'\leftarrow t}\right)\\
&\qquad\qquad\quad
 + \lambda_t\, \left\|\hat{\mathbf{t}}^{\mathrm{aux}}_{t'\leftarrow t}-\mathbf{t}_{t'\leftarrow t}\right\|_1
\Big).
\end{aligned}
\label{eq:aux-loss}
\end{equation}
\begin{equation}
\mathcal{L} = \mathcal{L}_{\mathrm{rel}} + \mathcal{L}_{\mathrm{aux}}.
\label{eq:total-loss}
\end{equation}
The loss encourages camera tokens to encode locally consistent motion early in the network, while the final global
stage enforces long-range trajectory consistency.

\section{Experiments}

We evaluate feedforward monocular visual odometry on long driving trajectories, with details described below.

\subsection{Datasets}
We conduct evaluation on two standard benchmarks following established protocols (see~\cite{deng2025vggtlong}), and one large-scale proprietary dataset:
\paragraph{Waymo Open Dataset (WOD)~\cite{sun2020waymo}}
The evaluation set contains 9 urban driving scenes. We do not use WOD for training our pose decoder.

\paragraph{KITTI Odometry~\cite{geiger2012kitti}}
This evaluation dataset contains 11 long driving scenes, with trajectory lengths spanning from 393.65 to 5067.23 meters. We do not use KITTI for training our pose coder.

\paragraph{Proprietary Driving Dataset (PDD)}
To further validate robustness, we evaluate on a large-scale proprietary dataset. This collection comprises 13,140 sequences, each spanning 100 frames, encompassing diverse weather conditions, lighting (day/night), and varying ego-speeds. We adopt a 90/10 train-test split, resulting in 11,826 segments for training and 1,314 for evaluation.

\subsection{Baselines}
On WOD and PDD, we evaluate feedforward geometry baselines VGGT~\cite{wang2025vggt} and
$\pi^3$~\cite{wang2025pi3}, and also report $\pi^3$ after fine-tuning on PDD
(dubbed \pithree-finetuned). For KITTI, we compare against long-horizon
feedforward pipelines, including VGGT-Long~\cite{deng2025vggtlong},
\pilong~\cite{deng2025pilong}, and Depth Anything 3 in streaming mode
(\dastream)~\cite{lin2025da3}. These methods use feedforward geometry followed by
trajectory-level post-processing; our \ours-Long uses the same post-optimization
setup for fair comparison. We further include recent monocular RGB VO/SLAM
results reported by prior work, including SCE-SLAM~\cite{wu2026sceslam},
VGGT-Motion~\cite{xiong2026vggtmotion}, Scal3R~\cite{xie2026scal3r},
HyVGGT-VO~\cite{pan2026hyvggtvo}, LingBot-Map~\cite{chen2026geometric},
MVOFormer~\cite{li2026mvoformer}, and
PoseFM~\cite{kuczkowski2026posefm}.

\subsection{Metrics}
\textbf{Absolute Trajectory Error (ATE, RMSE).} We evaluate global trajectory accuracy using
the root mean square error (RMSE) of Euclidean distances between predicted and ground-truth camera
positions after per-sequence Sim(3) alignment \cite{umeyama2002least}, and report ATE in meters.

\textbf{Relative Pose Error (RPE, $\Delta$=1 frame).} We evaluate local drift using relative pose
error between consecutive frames ($\Delta$=1), and report translation RMSE (meters) and rotation
RMSE (degrees). For WOD and PDD, we report the mean over sequences.

\begin{table}[ht]
\centering
\caption{
\textbf{WOD eval results.}
$^\dagger$: ATE-only results reported by prior work.}
\label{tab:wod-eval}
\setlength{\tabcolsep}{3pt}
\resizebox{\columnwidth}{!}{%
\begin{tabular}{lccc}
\toprule
Method & ATE (m) $\downarrow$ & RPE$_t$ (m) $\downarrow$ & RPE$_r$ (deg) $\downarrow$ \\%
\midrule%
DROID-SLAM~\cite{teed2021droidslam}$^\dagger$ & 4.396 & -- & -- \\
MASt3R-SLAM~\cite{murai2024mast3rslam}$^\dagger$  & 5.560 & -- & -- \\
CUT3R~\cite{wang2025continuous}$^\dagger$  & 9.872 & -- & -- \\
VGGT-Long~\cite{deng2025vggtlong} $^\dagger$ & 1.996 & -- & -- \\
SCE-SLAM~\cite{wu2026sceslam}$^\dagger$ & 0.915 & -- & -- \\
VGGT-Motion~\cite{xiong2026vggtmotion}$^\dagger$ & 1.587 & -- & -- \\
Scal3R~\cite{xie2026scal3r}$^\dagger$ & 1.520 & -- & -- \\
\midrule
VGGT~\cite{wang2025vggt}             & 1.359 & 0.299 & 0.174 \\
\pithree~\cite{wang2025pi3}              & 1.150 & 0.156 & \best{0.113} \\
\pithree-finetuned    & 1.170 & 0.332 & 0.125 \\
\midrule
\textbf{\ours}~w/o augmentation         & \best{0.714} & \best{0.118} & \secondbest{0.124} \\
\textbf{\ours} & \secondbest{0.862} & \secondbest{0.114} & {0.128} \\
\bottomrule
\end{tabular}
}
\end{table}

\begin{table}[h]
\centering
\caption{
\textbf{Proprietary Driving Dataset (PDD) eval results.}
}
\label{tab:internal-eval}
\setlength{\tabcolsep}{3pt}
\resizebox{\columnwidth}{!}{%
\begin{tabular}{lccc}
\toprule
Method & ATE (m) $\downarrow$ & RPE$_t$ (m) $\downarrow$ & RPE$_r$ (deg) $\downarrow$ \\%
\midrule
VGGT~\cite{wang2025vggt}             & 10.194 & 2.628 & 0.483 \\
\pithree~\cite{wang2025pi3}              & 7.928  & 1.773 & 0.249 \\
\pithree-finetuned    & 5.720  & 2.713 & 0.269 \\
\midrule
\textbf{\ours}~w/o augmentation      & \secondbest{2.779} & \best{0.619} & \best{0.184} \\
\textbf{\ours}  & \best{2.494} & \secondbest{0.621} & \secondbest{0.195} \\ 
\bottomrule
\end{tabular}
}
\end{table}

\newcommand{\pddfigpath}{images/pdd_traj_plot/}
\newcommand{\wodfigpath}{images/wod_traj_plot/}
\newcommand{\trajimgscale}{0.19}
\begin{figure*}[h] 
    \centering
    \begin{subfigure}{\trajimgscale\textwidth}
        \includegraphics[width=\textwidth]{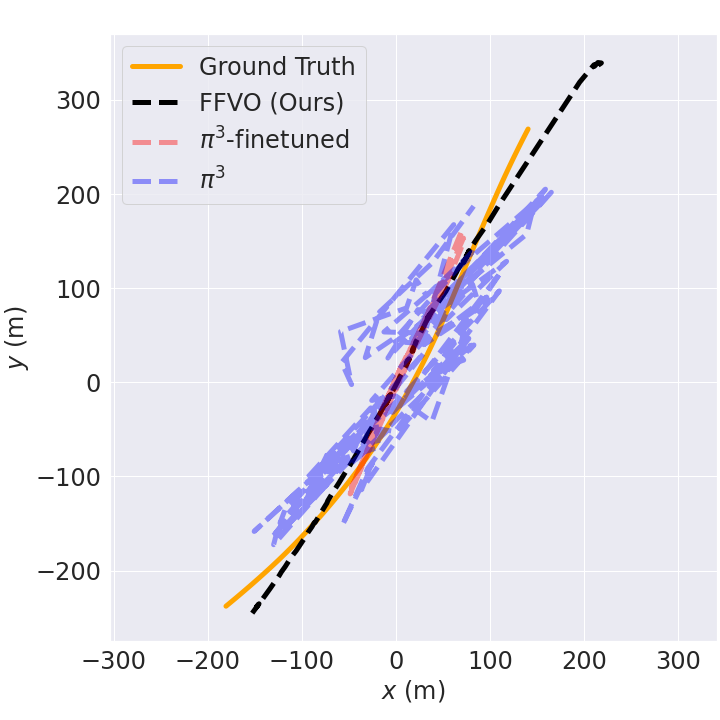}%
    \end{subfigure}
    \hfill
    \begin{subfigure}{\trajimgscale\textwidth}
        \includegraphics[width=\textwidth]{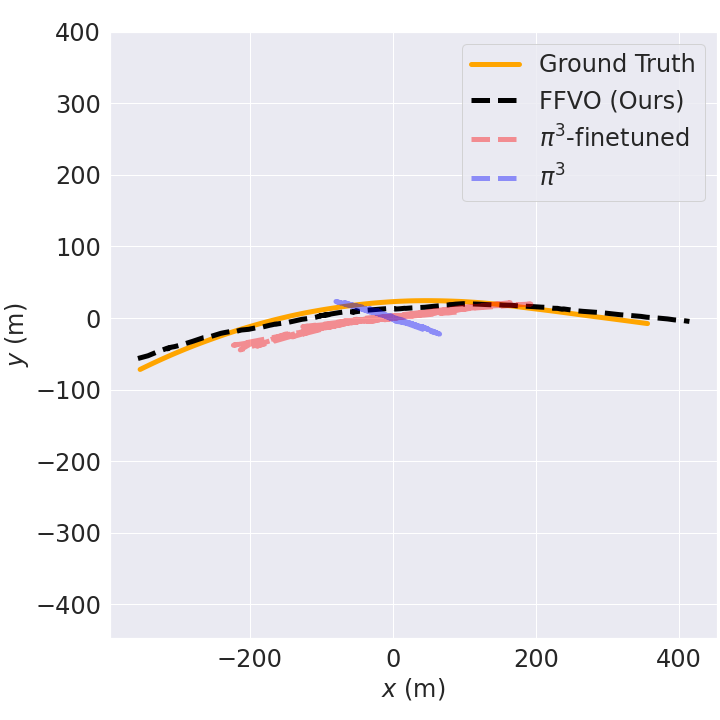}%
    \end{subfigure}
    \hfill
    \begin{subfigure}{\trajimgscale\textwidth}
        \includegraphics[width=\textwidth]{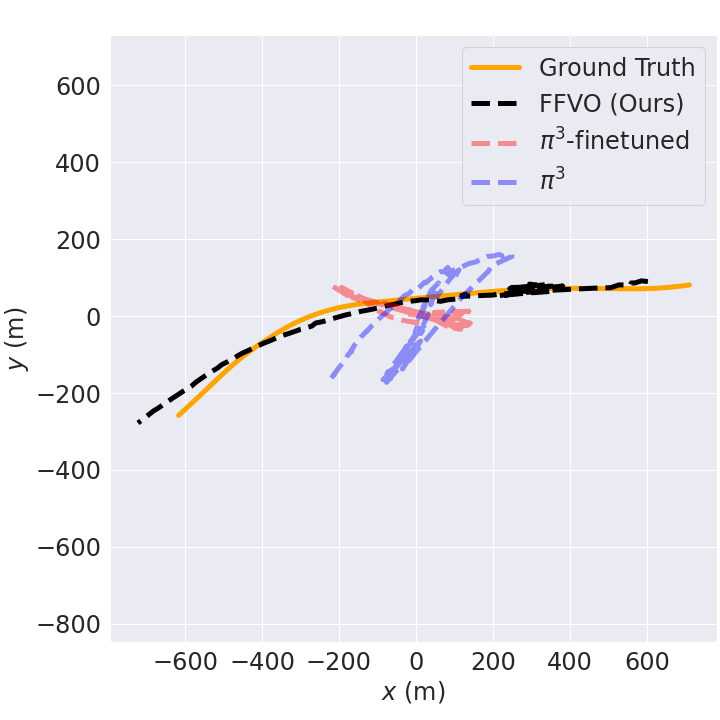}%
    \end{subfigure}
    \hfill
    \begin{subfigure}{\trajimgscale\textwidth}
        \includegraphics[width=\textwidth]{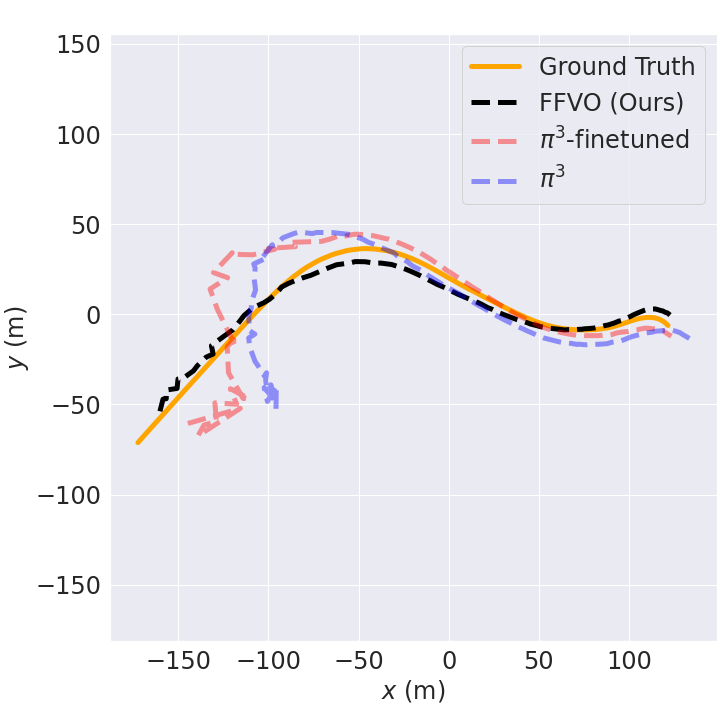}%
    \end{subfigure}
    \caption*{Sequences from the PDD test set with fast, long trajectories.}
    \begin{subfigure}{\trajimgscale\textwidth}
        \includegraphics[width=\textwidth]{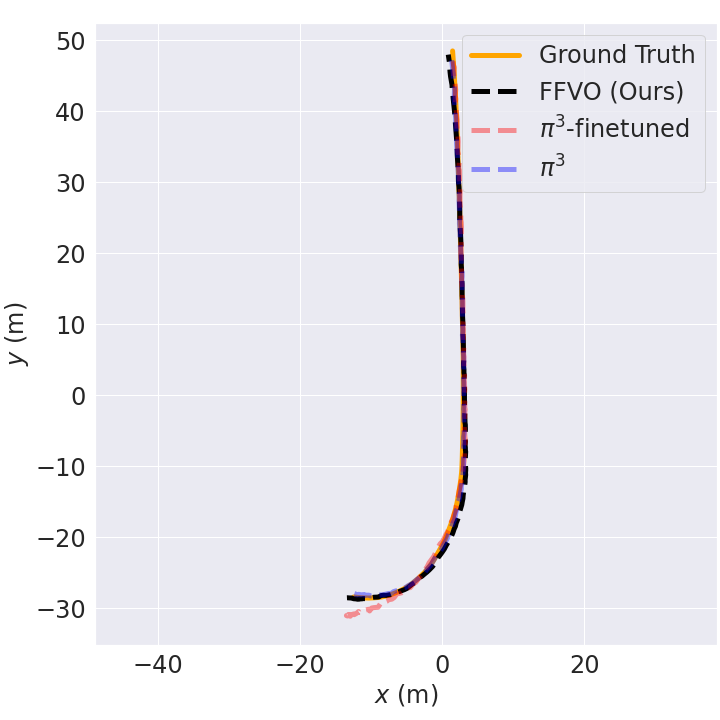}
        \caption*{WOD-40584103}
    \end{subfigure}
    \hfill
    \begin{subfigure}{\trajimgscale\textwidth}
        \includegraphics[width=\textwidth]{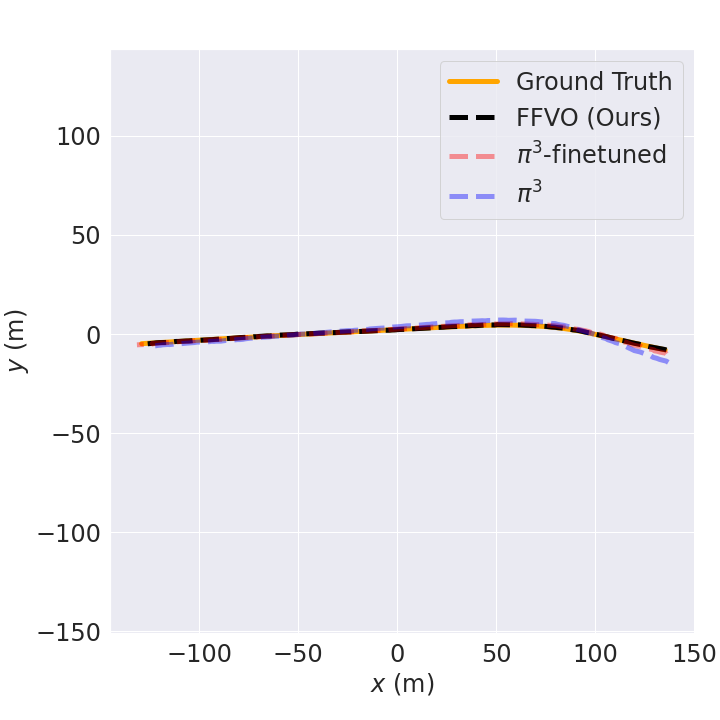}
        \caption*{WOD-46041731}
    \end{subfigure}
    \hfill
    \begin{subfigure}{\trajimgscale\textwidth}
        \includegraphics[width=\textwidth]{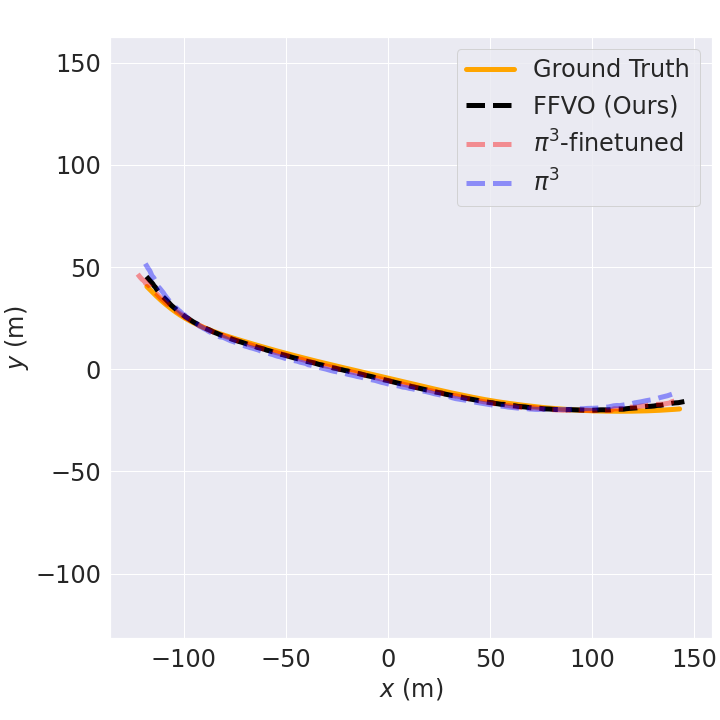}
        \caption*{WOD-37115986}
    \end{subfigure}
    \hfill
    \begin{subfigure}{\trajimgscale\textwidth}
        \includegraphics[width=\textwidth]{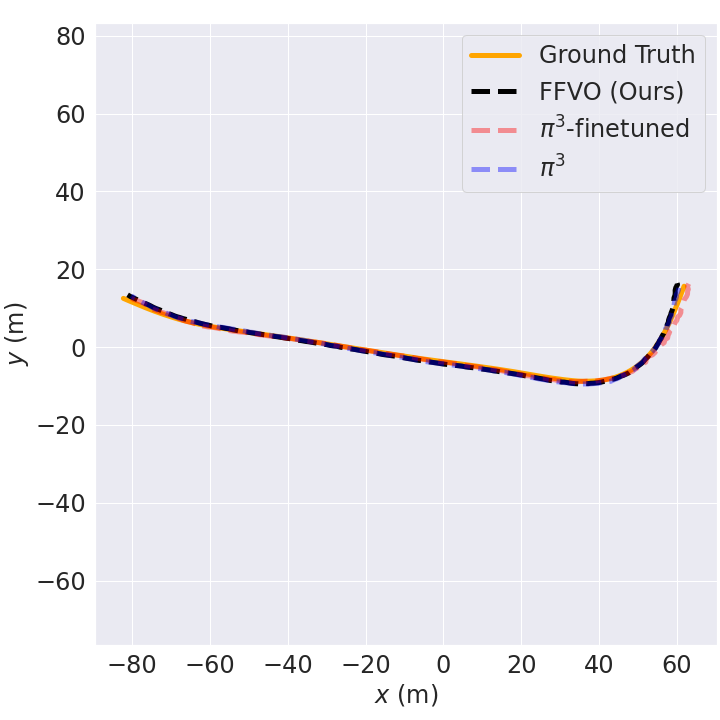}
        \caption*{WOD-16345319}
    \end{subfigure}
    \caption{
    \textbf{Qualitative long-horizon odometry results.}
    We visualize predicted trajectories versus ground truth and show how local-to-global decoding reduces drift over long driving sequences.}
    \label{fig:qual-traj}
\end{figure*}

\newcommand{\kittiqualcloudpath}{images/kitti_qual_cloud/cropped/}
\newcommand{\kittiqualimagescale}{0.4}
\begin{figure}[t]
    \centering
    \hfill%
    \begin{subfigure}[t]{\kittiqualimagescale\linewidth}
        \centering
        \includegraphics[width=0.65\textwidth]{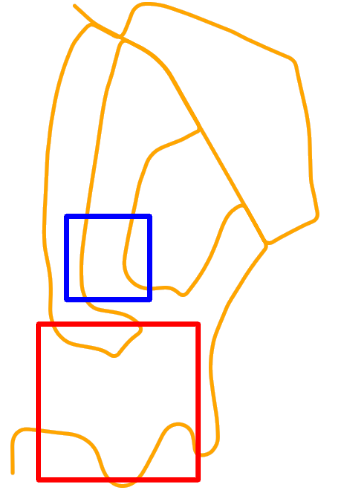}
        \caption*{GT}
        \includegraphics[width=0.65\textwidth]{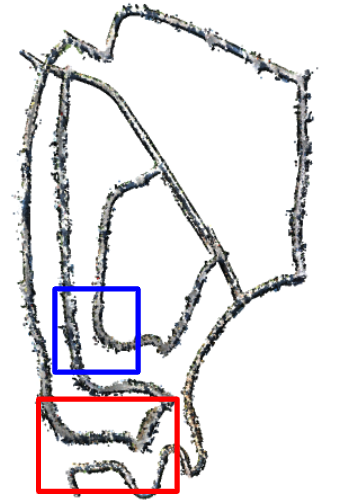}
        \caption*{DA3-Streaming~\cite{lin2025da3}}
    \end{subfigure}
    \hfill%
    \begin{subfigure}[t]{\kittiqualimagescale\linewidth}
        \centering
        \includegraphics[width=0.65\textwidth]{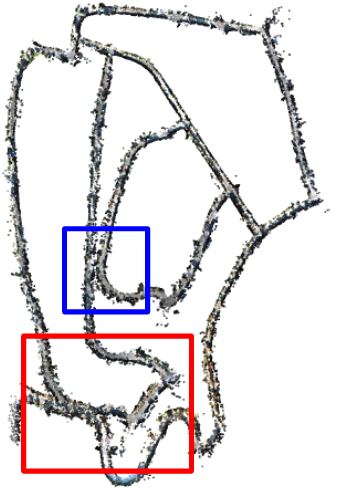}
        \caption*{$\pi^3$-Long~\cite{deng2025pilong}}
        \includegraphics[width=0.65\textwidth]{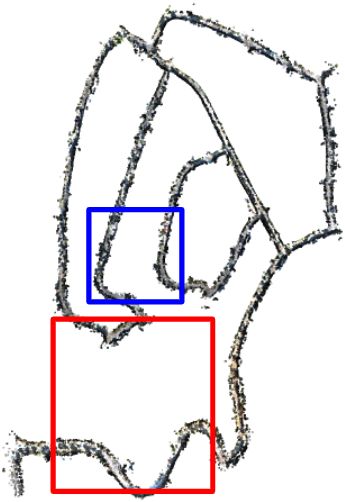}
        \caption*{\ours-Long (Ours)}
    \end{subfigure}
    \hfill%
    \caption{
    \textbf{Qualitative comparison on KITTI Odometry Seq.~02.}
    The blue and red boxes highlight local trajectory geometry: $\pi^3$-Long shows a local self-collision in the blue box, and both $\pi^3$-Long and DA3-Streaming produce overly-close/merged curves in the red box, while Ours-Long better preserves the separation and overall shape compared to GT.}
    \label{fig:kitti-qual-cloud}
\end{figure}

\newcommand{\failurecaseimagepath}{images/limitation/}
\newcommand{\failurecaseimagescale}{0.29}

\begin{figure}[t]
\centering
    \begin{subfigure}{\failurecaseimagescale\linewidth}
        \includegraphics[width=\textwidth]{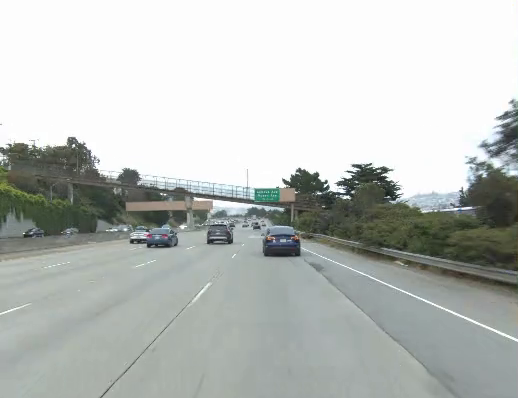}\\[0.4ex]%
        \includegraphics[width=\textwidth]{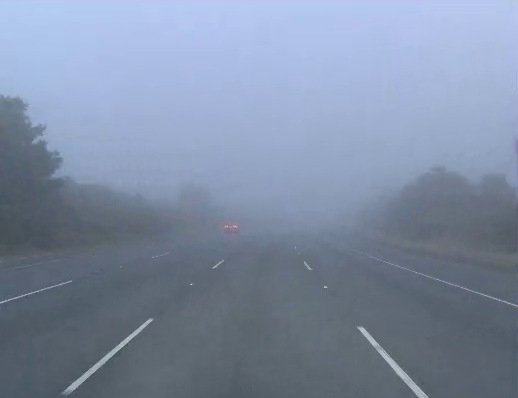}\\[0.4ex]%
        \includegraphics[width=\textwidth]{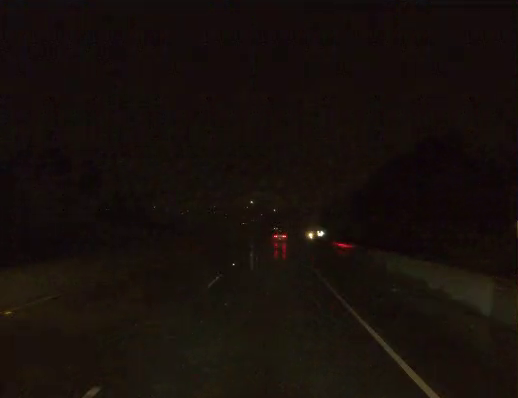}%
    \caption*{Reference}%
    \label{fig:failure-input}%
    \end{subfigure}
    \hfill
    \begin{subfigure}{\failurecaseimagescale\linewidth}
        \includegraphics[width=\textwidth]{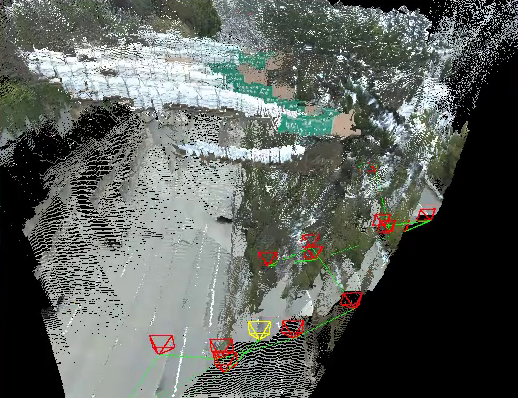}\\[0.4ex]%
        \includegraphics[width=\textwidth]{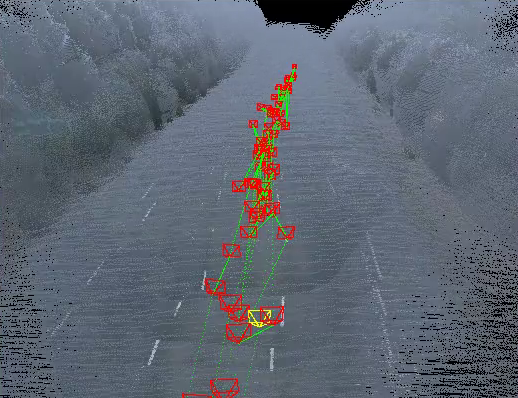}\\[0.4ex]%
        \includegraphics[width=\textwidth]{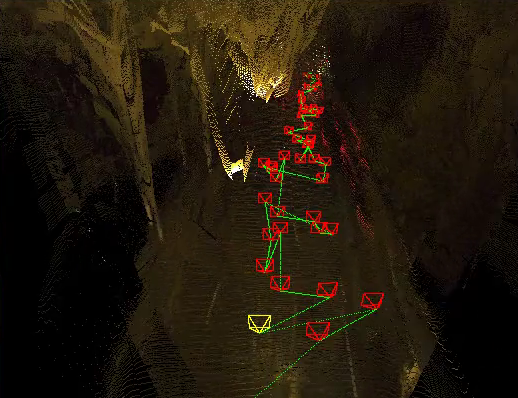}%
    \caption*{\pithree \cite{wang2025pi3}}%
    \label{fig:failure-pi3}%
    \end{subfigure}
    \hfill
    \begin{subfigure}{\failurecaseimagescale\linewidth}
        \includegraphics[width=\textwidth]{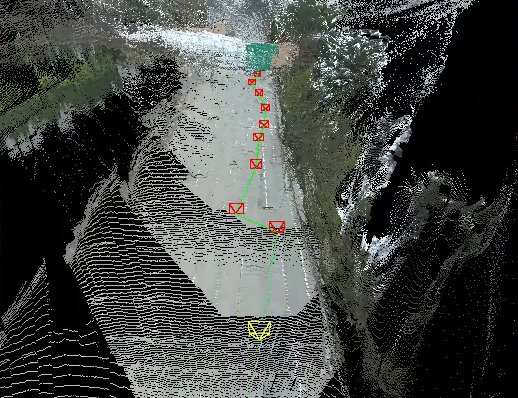}\\[0.4ex]%
        \includegraphics[width=\textwidth]{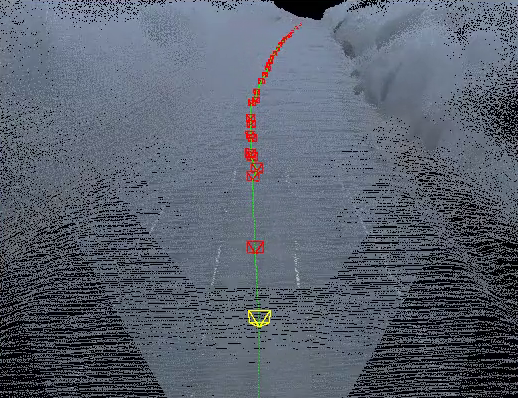}\\[0.4ex]%
        \includegraphics[width=\textwidth]{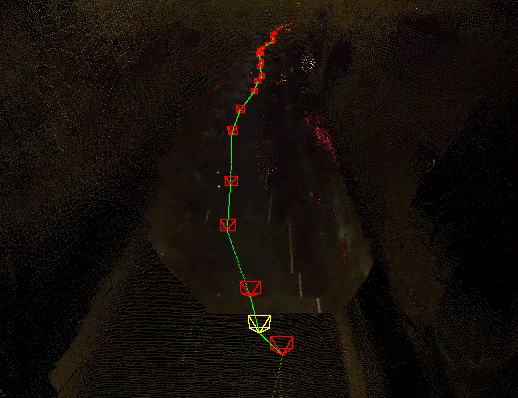}%
    \caption*{\ourspost}%
    \label{fig:failure-ours}%
    \end{subfigure}
\caption{
\textbf{Failure modes.}
Our model shows similar patterns of temporal instability and inaccuracy as other feedforward methods under degraded conditions: (top) high-speed motion exceeding 55 mph, (center) extreme weather, and (bottom) low-light environments.
}
\label{fig:failure-cases}
\end{figure}

\begin{table*}[h]
\centering
\vspace{0.25em}
\caption{
\textbf{KITTI Odometry eval results}.
Top-2 highlighting excludes classical methods. $^\dagger$: from \cite{deng2025vggtlong}. $^\ddagger$: per-sequence values are reported where available.}
\label{tab:kitti-eval}
\scriptsize
\setlength{\tabcolsep}{3.2pt}
\renewcommand{\arraystretch}{1.04}
\resizebox{\textwidth}{!}{
\begin{tabular}{c|l|c|c c c c c c c c c c c}
\toprule
 & Methods & Avg. & \textbf{00} & \textbf{01} & \textbf{02} & \textbf{03} & \textbf{04} & \textbf{05} & \textbf{06} & \textbf{07} & \textbf{08} & \textbf{09} & \textbf{10} \\
\midrule
\multirow{3}{*}{\rotatebox[origin=c]{90}{Classical}} & ORB-SLAM2 w/o LC~\cite{mur2017orb}$^\dagger$ & 69.727 & 40.65 & 502.20 & 47.82 & 0.94 & 1.30 & 29.95 & 40.82 & 16.04 & 43.09 & 38.77 & 5.42 \\
& ORB-SLAM2 w/ LC~\cite{mur2017orb}$^\dagger$ & 54.816 & 6.03 & 508.34 & 14.76 & 1.02 & 1.57 & 4.04 & 11.16 & 2.19 & 38.85 & 8.39 & 6.63 \\
& LDSO~\cite{gao2018ldso}$^\dagger$ & 22.425 & 9.32 & 11.68 & 31.98 & 2.85 & 1.22 & 5.10 & 13.55 & 2.96 & 129.02 & 21.64 & 17.36 \\
\midrule
\multirow{17}{*}{\rotatebox[origin=c]{90}{Learning Based}} & DROID-VO~\cite{teed2021droidslam}$^\dagger$ & 54.188 & 98.43 & 84.20 & 108.80 & 2.58 & 0.93 & 59.27 & 64.40 & 24.20 & 64.55 & 71.80 & 16.91 \\
& DROID-SLAM~\cite{teed2021droidslam}$^\dagger$ & 100.278 & 92.10 & 344.60 & 107.61 & 2.38 & 1.00 & 118.50 & 62.47 & 21.78 & 161.60 & 72.32 & 118.70 \\
& DPVO~\cite{teed2023deep}$^\dagger$ & 53.609 & 113.21 & 12.69 & 123.40 & 2.09 & 0.68 & 58.96 & 54.78 & 19.26 & 115.90 & 75.10 & 13.63 \\
& DPV-SLAM~\cite{lipson2024deep}$^\dagger$ & 53.034 & 112.80 & 11.50 & 123.53 & 2.50 & 0.81 & 57.80 & 54.86 & 18.77 & 110.49 & 76.66 & 13.65 \\
& DPV-SLAM++~\cite{lipson2024deep}$^\dagger$ & 25.749 & 8.30 & 11.86 & 39.64 & 2.50 & 0.78 & 5.74 & 11.60 & \best{1.52} & 110.90 & 76.70 & 13.70 \\
& SCE-SLAM w/o LC~\cite{wu2026sceslam} & 25.790 & 53.31 & \best{9.63} & 62.32 & 2.05 & \best{0.54} & 31.34 & 26.37 & 11.65 & 43.67 & 30.19 & 12.57 \\
& SCE-SLAM w/ LC~\cite{wu2026sceslam} & \secondbest{14.070} & 8.11 & \secondbest{9.82} & \secondbest{31.04} & 2.40 & \secondbest{0.56} & 5.40 & 11.17 & \secondbest{1.95} & 41.21 & 30.38 & 12.66 \\
& VGGT-Long~\cite{deng2025vggtlong} & 25.600 & 16.13 & 53.43 & 51.98 & 4.37 & 2.15 & 12.69 & 11.33 & 3.60 & 70.29 & 34.55 & 21.05 \\
& HyVGGT-VO (optimized)~\cite{pan2026hyvggtvo} & 59.642 & 114.29 & 47.22 & 142.01 & 2.19 & 0.60 & 68.14 & 61.21 & 13.36 & 115.14 & 76.26 & 15.65 \\
& VGGT-Motion~\cite{xiong2026vggtmotion} & 24.170 & 6.79 & 83.29 & 66.48 & 7.08 & 2.78 & 7.63 & 4.06 & 3.72 & 39.93 & 28.77 & 15.35 \\
& LingBot-Map~\cite{chen2026geometric} & 24.046 & -- & -- & -- & -- & -- & -- & -- & -- & -- & -- & -- \\
& Scal3R~\cite{xie2026scal3r}$^\ddagger$ & 14.550 & \best{4.30} & 45.29 & -- & -- & -- & \best{3.30} & -- & 2.03 & -- & 12.32 & -- \\
& MVOFormer~\cite{li2026mvoformer} & 19.610 & 22.37 & 53.70 & 67.15 & 3.13 & 1.63 & 9.91 & 14.24 & 6.52 & \best{19.39} & 13.84 & \best{3.81} \\
& PoseFM+PWC~\cite{kuczkowski2026posefm} & 38.740 & 59.66 & 123.48 & 63.36 & 6.42 & 3.55 & 49.74 & 32.42 & 4.56 & 24.86 & 45.36 & 12.76 \\
& \pithree-Long~\cite{deng2025pilong} & 21.180 & 5.55 & 114.83 & 50.29 & \best{1.63} & 1.11 & \secondbest{3.48} & \secondbest{2.88} & 3.92 & \secondbest{24.25} & \best{7.38} & 17.61 \\
& DA3-Streaming~\cite{lin2025da3} & 18.630 & \secondbest{4.48} & 100.77 & 33.41 & 3.58 & 2.39 & 3.95 & 7.59 & 2.09 & 31.20 & 8.06 & \secondbest{7.44} \\
\cmidrule(lr){2-14}
& \textbf{\ours-Long} & \best{12.960} & 4.82 & 49.47 & \best{30.25} & \secondbest{2.00} & 0.80 & 3.94 & \best{2.25} & 2.34 & 25.49 & \secondbest{7.56} & 13.67 \\
\bottomrule
\end{tabular}
}
\end{table*}

\newcommand{\kittipath}{images/kitti_traj_plot/}
\newcommand{\kittitrajimgscale}{0.132}

\begin{figure*}[t]
    \centering
    \begin{subfigure}{\kittitrajimgscale\linewidth}
        \includegraphics[width=\linewidth]{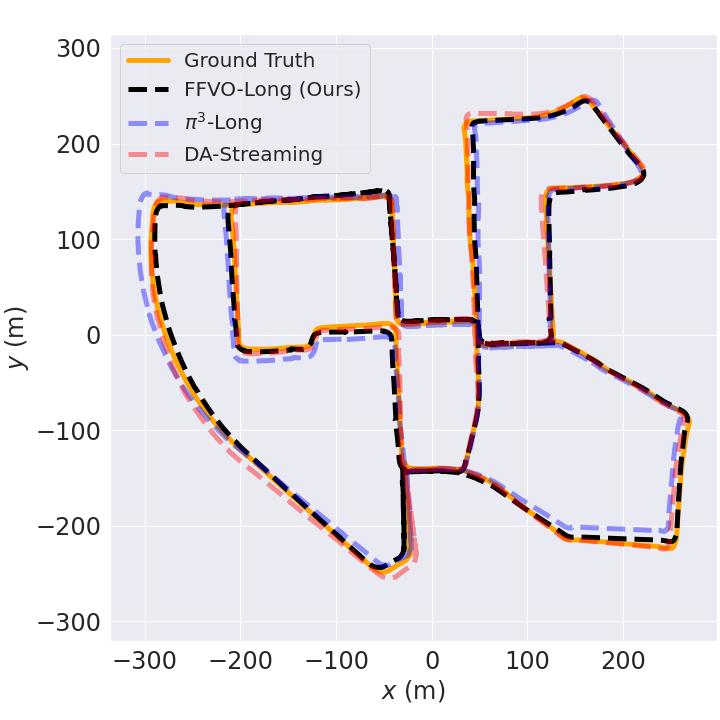}
        \caption*{KITTI-00}
    \end{subfigure}
    \begin{subfigure}{\kittitrajimgscale\linewidth}
        \includegraphics[width=\linewidth]{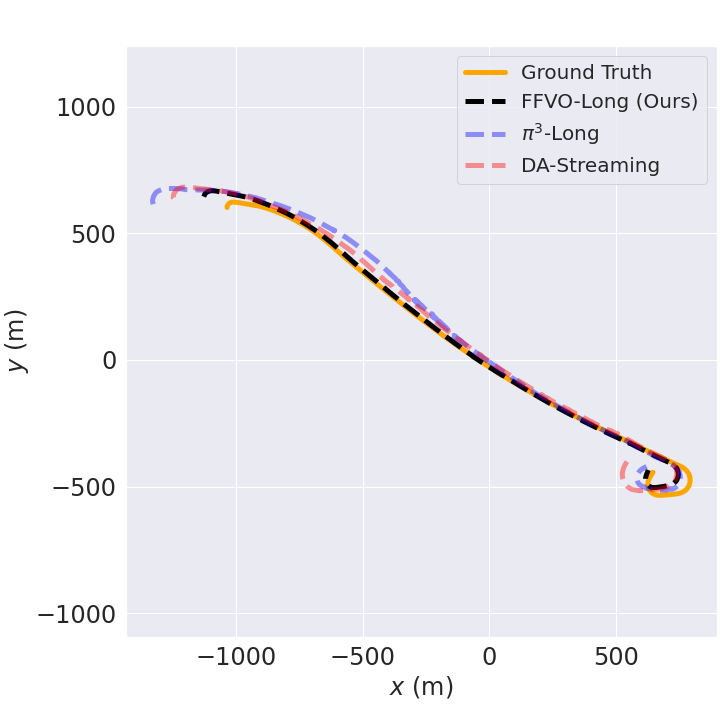}
        \caption*{KITTI-01}
    \end{subfigure}
    \begin{subfigure}{\kittitrajimgscale\linewidth}
        \includegraphics[width=\linewidth]{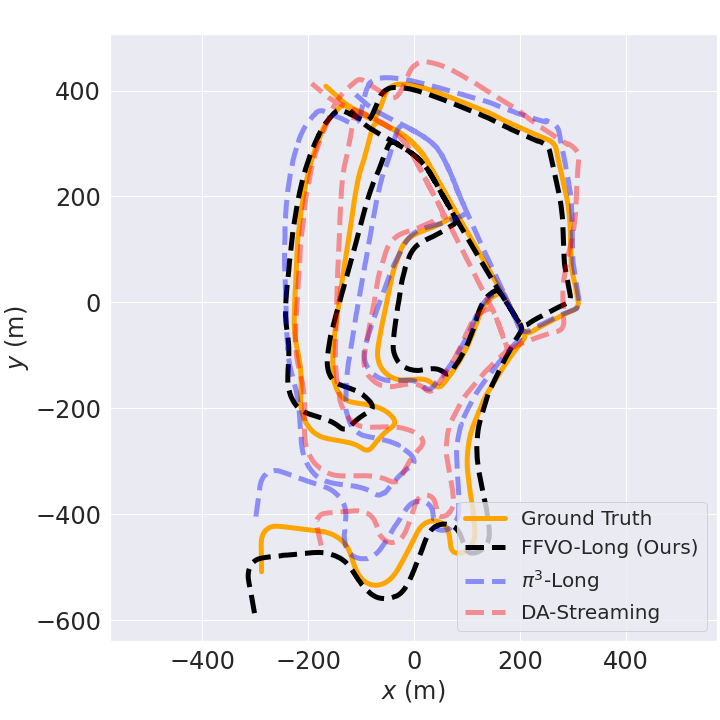}
        \caption*{KITTI-02}
    \end{subfigure}
    \begin{subfigure}{\kittitrajimgscale\linewidth}
        \includegraphics[width=\linewidth]{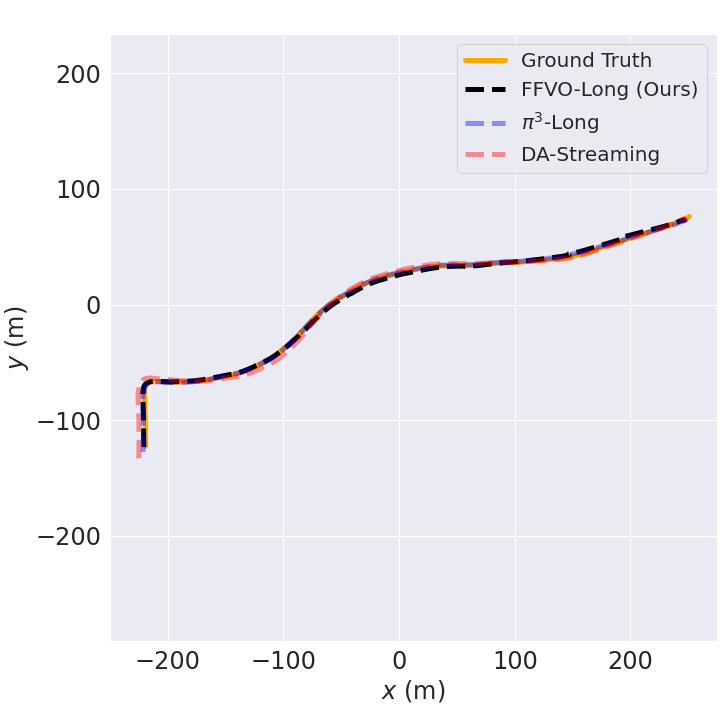}
        \caption*{KITTI-03}
    \end{subfigure}
    \begin{subfigure}{\kittitrajimgscale\linewidth}
        \includegraphics[width=\linewidth]{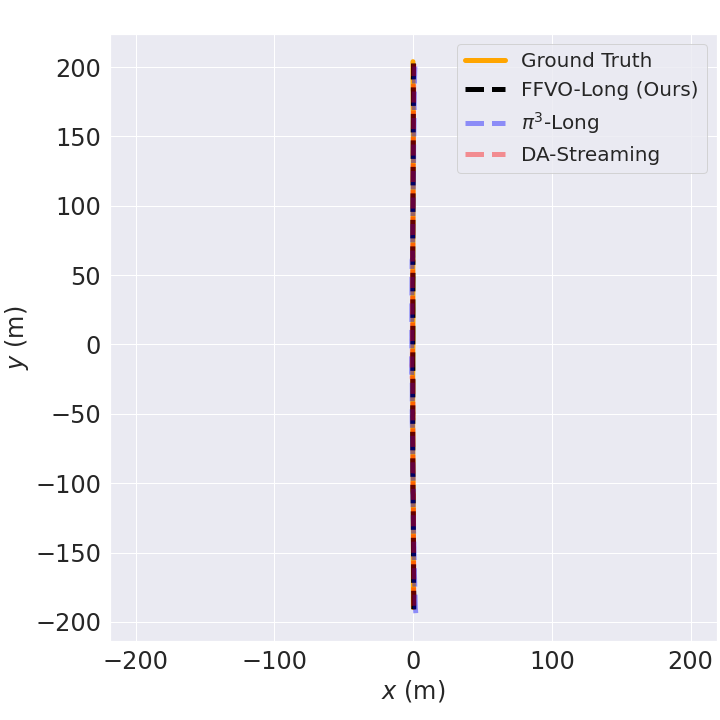}
        \caption*{KITTI-04}
    \end{subfigure}
    \begin{subfigure}{\kittitrajimgscale\linewidth}
        \includegraphics[width=\linewidth]{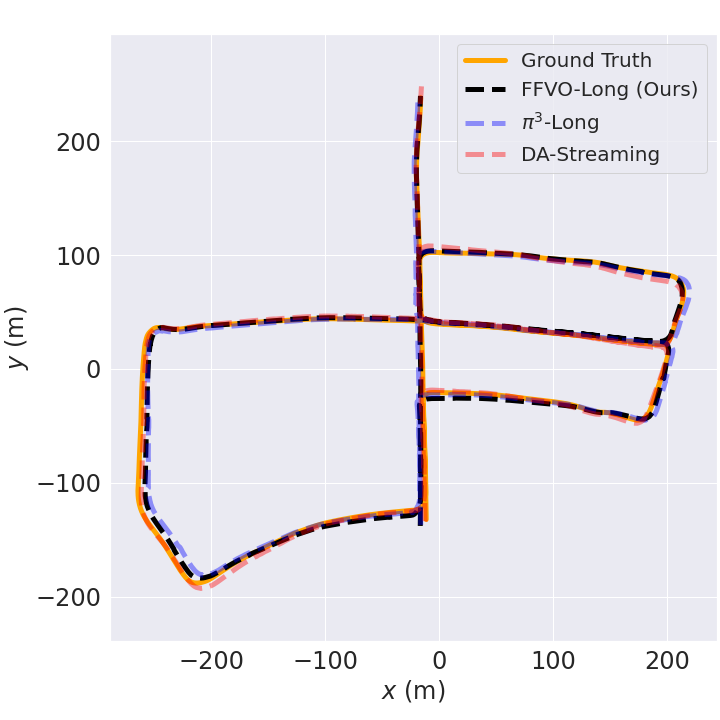}
        \caption*{KITTI-05}
    \end{subfigure}
    \\
    \begin{subfigure}{\kittitrajimgscale\linewidth}
        \includegraphics[width=\linewidth]{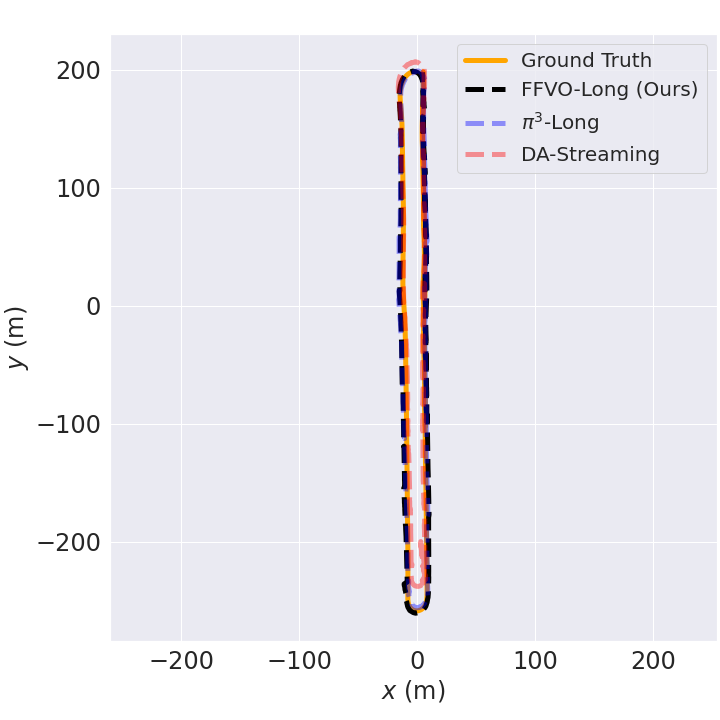}
        \caption*{KITTI-06}
    \end{subfigure}
    \begin{subfigure}{\kittitrajimgscale\linewidth}
        \includegraphics[width=\linewidth]{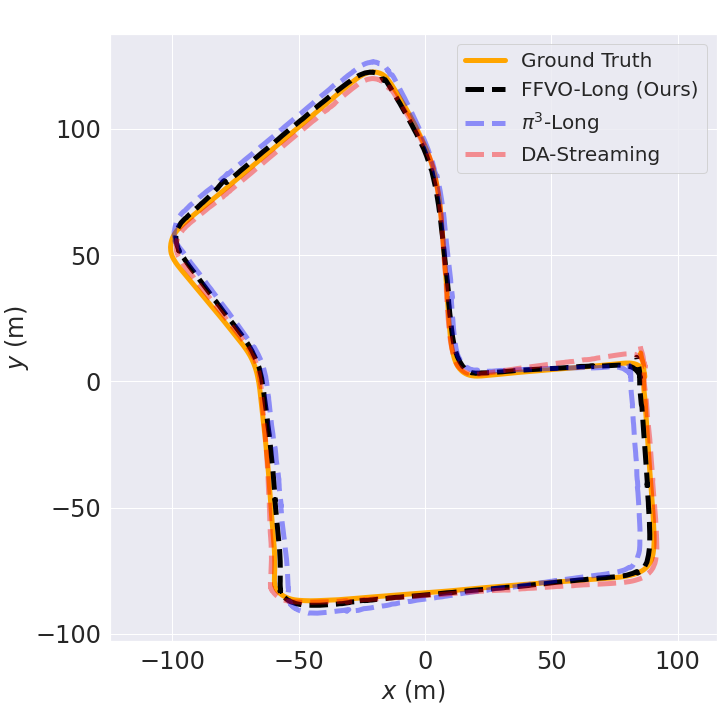}
        \caption*{KITTI-07}
    \end{subfigure}
    \begin{subfigure}{\kittitrajimgscale\linewidth}
        \includegraphics[width=\linewidth]{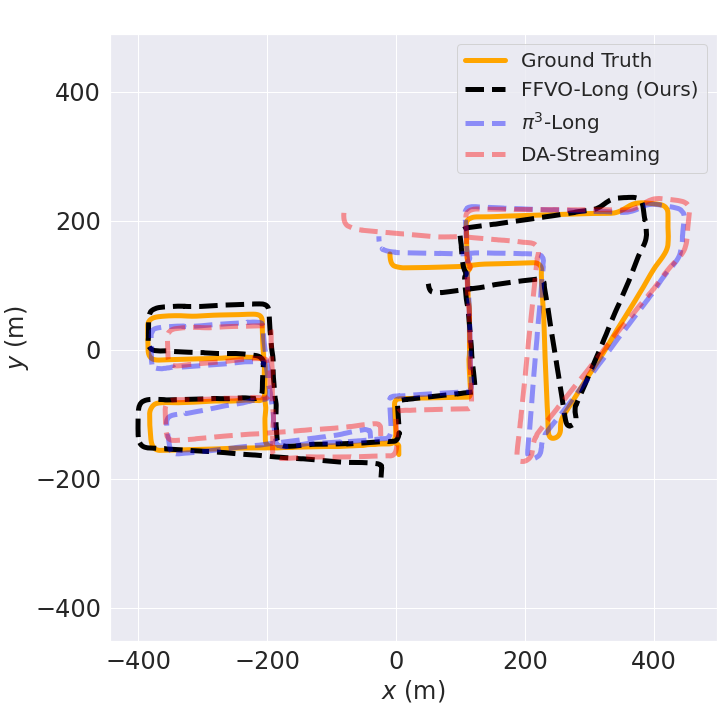}
        \caption*{KITTI-08}
    \end{subfigure}
    \begin{subfigure}{\kittitrajimgscale\linewidth}
        \includegraphics[width=\linewidth]{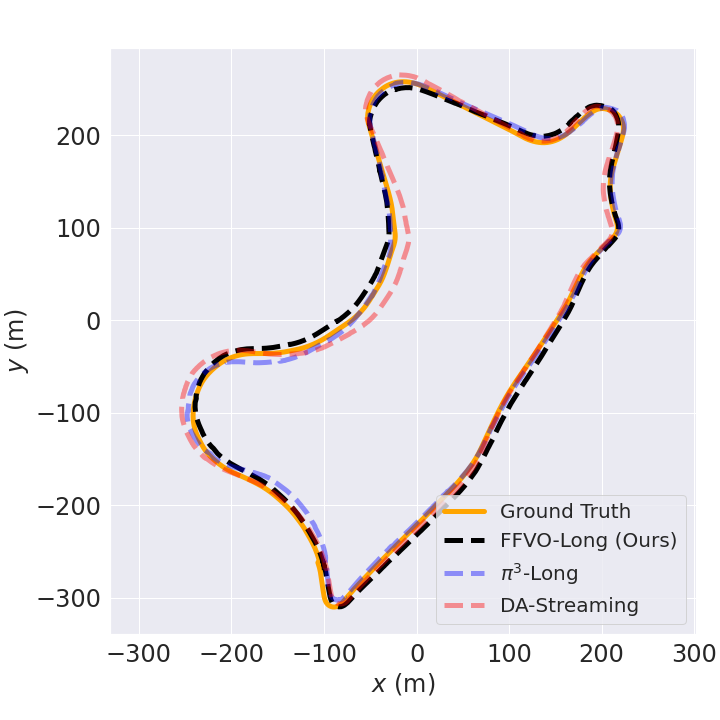}
        \caption*{KITTI-09}
    \end{subfigure}
    \begin{subfigure}{\kittitrajimgscale\linewidth}
        \includegraphics[width=\linewidth]{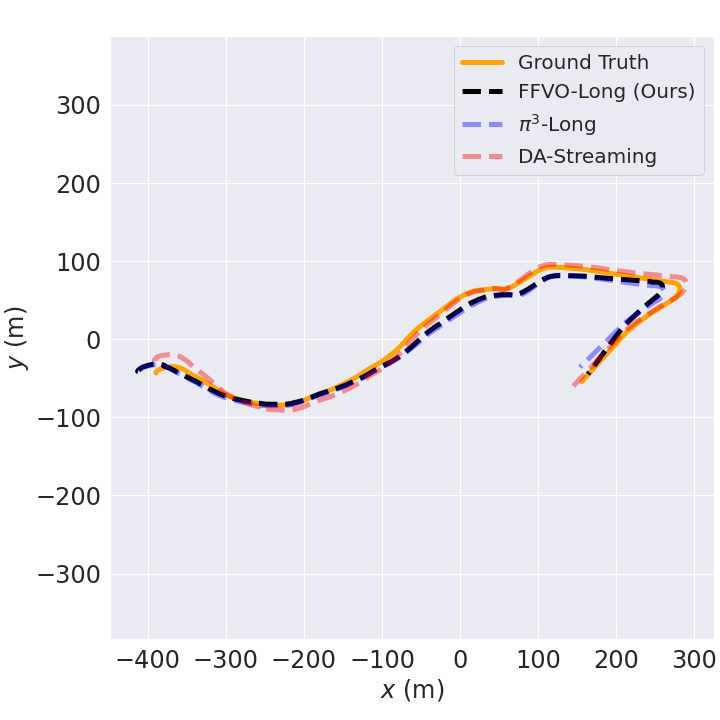}
        \caption*{KITTI-10}
    \end{subfigure}
    \caption{
    \textbf{Trajectory visualization on the KITTI Odometry Benchmark.}
    Predictions and ground truth for sequences 00--10; all methods use 120-frame alignment chunks~\cite{deng2025vggtlong}.
    }
    \label{fig:kitti-all-traj}
\end{figure*}

\subsection{Implementation Details}
We start from a pretrained $\pi^3$ backbone \cite{wang2025pi3} and freeze the feature aggregation
and point-map modules, training only the pose decoding branch (described in \secref{sec:method}) from scratch.
We use $M{=}14$ camera tokens per frame and a local temporal window of $w{=}20$ frames, selected empirically based on our GPU memory constraints. The pose
decoder comprises two local-window temporal stages, followed by a global integration stage and two
frame self-attention layers; the auxiliary local-trajectory head is attached after the second local
stage (\secref{sec:method}).
We train on PDD only, with sequence length $T{=}100$ frames (matching the segment
length at inference). We use a total batch size of 64 on 8 NVIDIA B200 GPUs with gradient
accumulation, and follow $\pi^3$'s optimization hyperparameters. For pose supervision, we set
$\lambda_R{=}1$ and $\lambda_t{=}100$ (rotation and translation weights, respectively), following
$\pi^3$. Our
\emph{augmentation} setting adds a wide-FoV (85$^\circ$ horizontal) video stream created by stitching
the front, front-left, and front-right cameras, in addition to the standard front camera stream; we
report models trained with both streams as ``\ours'' and models trained without augmentation as
``\ours\ w/o augmentation'' in \tabsref{tab:wod-eval}--\ref{tab:internal-eval}. For evaluation on PDD,
we report results using the front camera only.
Training takes approximately 2 days for the low-resolution stage and 3 days for the high-resolution
stage on 8 NVIDIA B200 GPUs.
Unless otherwise specified, all quantitative and qualitative experiments use an input width of
518 pixels; only the ablation study uses 224 pixels for faster iteration.

\subsection{Quantitative Results}
\tabsref{tab:wod-eval} and \ref{tab:internal-eval} report ATE and RPE on WOD and PDD, respectively.
\tabref{tab:kitti-eval} reports per-sequence ATE on KITTI Odometry sequences 00--10.
For completeness, \tabref{tab:wod-eval} also includes additional ATE-only WOD baselines reported by
prior work~\cite{deng2025vggtlong,wu2026sceslam,xiong2026vggtmotion,xie2026scal3r}; those entries do not provide RPE.
\tabref{tab:kitti-eval} likewise includes additional KITTI entries from Table 1
of~\cite{deng2025vggtlong}, excluding OOM and TL cases where tracking does not complete.

On WOD (\tabref{tab:wod-eval}), \ours\ outperforms pretrained and fine-tuned feedforward methods
on ATE and RPE$_t$, and remains competitive on RPE$_r$~\cite{lin2025da3,deng2025vggtlong,deng2025pilong}.
Compared with \pithree-finetuned, the gains indicate that the improvement is not only due to training data, but also to the architectural design of \ours\ pose decoder. We observe a performance drop in \pithree-finetuned. We attribute this to the domain gap between WOD and PDD; the latter's broader environmental conditions (e.g., weather, lighting) require \pithree-finetuned to generalize beyond WOD's distribution.
On PDD (\tabref{tab:internal-eval}), \ours\ significantly outperforms all baselines across all metrics,
demonstrating robust generalization across diverse driving conditions in this dataset. The mixed augmentation results in \tabsref{tab:wod-eval}--\ref{tab:internal-eval}
may reflect an FoV mismatch between the stitched $85^\circ$ views used for augmentation and the
standard front-camera views used for WOD and PDD evaluation, which can reduce specialization to the
evaluation setting.
On KITTI (\tabref{tab:kitti-eval}), \ours-Long achieves the best overall ATE, showing that our
pose decoder integrates effectively with the same post-optimization pipeline used by prior feedforward methods
and still outperforms them~\cite{lin2025da3,deng2025vggtlong,deng2025pilong}.

\subsection{Qualitative Results}
\figref{fig:qual-traj} shows qualitative trajectory comparisons on WOD and PDD.
On PDD, \ours\ remains closely aligned with GT over the full trajectory, while feedforward baselines
without our tailored pose decoder show noticeable divergence and instability \cite{wang2025pi3}.
This supports the robustness of our pose decoder under long horizons and challenging conditions.
On WOD, both \ours\ and the baselines align well with GT because the evaluated trajectories are
shorter and generally less challenging.

\figref{fig:kitti-qual-cloud} provides a qualitative comparison on a challenging sequence (Seq.~02) in KITTI Odometry.
In the blue box, $\pi^3$-Long~\cite{deng2025pilong} exhibits a local trajectory 
self-collision (two passes collide), while DA3-Streaming~\cite{lin2025da3} and 
\ours-Long avoid this failure mode. In the red box, both $\pi^3$-Long and 
DA3-Streaming produce two curves that become too close (partly merged) compared
to GT, whereas \ours-Long better preserves the separation and overall shape. In \figref{fig:kitti-all-traj}, \ours{} shows overall good alignments across all KITTI sequences.

\subsection{Ablations}
\tabref{tab:ablation} studies the impact of (i) camera-token cross-attention, (ii) intermediate
auxiliary supervision, and (iii) restricting temporal attention to local windows. We train and evaluate all variants, including \pithree-finetuned, on \emph{PDD} without augmentation under the low-resolution setting
(224-pixel input width), which enables faster architecture iteration.

Starting from \pithree-finetuned, adding camera-token cross-attention improves all metrics, showing
that compact pose queries can effectively aggregate global context. Adding intermediate auxiliary
supervision yields a further, consistent gain, indicating that explicit supervision of intermediate
local motion improves pose decoding. Finally, local-window temporal attention provides the largest
overall improvement across metrics. This supports our design motivation: local temporal reasoning
reduces short-term ambiguity, while the subsequent global integration stage preserves long-range
consistency. Increasing the camera-token count for our default $M{=}14$ to
$M{=}24$ yields only limited gains.

\begin{table}[ht]
\setlength{\tabcolsep}{2pt}
\centering
\caption{\textbf{Ablation study on PDD.} All our proposed components contribute meaningfully to the final performance. \textcolor{improvegreen}{Green number} indicates relative improvement to \pithree-finetuned.}
\label{tab:ablation}
\resizebox{\columnwidth}{!}{%
\begin{tabular}{lccc}
\toprule
Method & ATE (m) $\downarrow$ & RPE$_t$ (m) $\downarrow$ & RPE$_r$ (deg) $\downarrow$ \\
\midrule
\pithree-finetuned & 15.989 \hphantom{\greenperc{26.35}} & 8.452 \hphantom{\greenperc{34.82}} & 0.596 \hphantom{\greenperc{36.41}} \\
+ camera tokens (cross-attn., $M{=14}$) & 11.776 \greenperc{26.35} & 5.509 \greenperc{34.82} & 0.379 \greenperc{36.41} \\
+ auxiliary supervision & 11.195 \greenperc{29.98} & 5.354 \greenperc{36.65} & 0.373 \greenperc{37.42} \\
+ local-window temporal attention & 7.070 \greenperc{55.78} & 1.090 \greenperc{87.10} & 0.289 \greenperc{51.51} \\
+ increase camera tokens to $M{=24}$ & 7.060 \greenperc{55.84} & 1.070 \greenperc{87.34} & 0.281 \greenperc{52.85} \\
\bottomrule
\end{tabular}%
}
\end{table}

\subsection{Limitations}
Despite strong overall performance, \ours\ shares feedforward monocular failure modes under severely
degraded conditions (\figref{fig:failure-cases}), including high-speed motion (over 55 mph),
extreme weather, and low light, which can cause temporal instability and larger errors. \ours-Long also
relies on segment-wise post-optimization for extremely long sequences because aggregating thousands
of frames exceeds memory. Sim(3)-aligned ATE and consecutive-frame RPE
may understate scale inconsistency and long-horizon drift; unaligned and multi-horizon evaluation,
end-to-end latency, throughput, and GPU memory comparisons with long-sequence baselines remain future
work. Finally, we focus on monocular front-camera data; multi-camera surround views could
improve robustness and accuracy.

\clearpage
{
\small
\bibliographystyle{ieeenat_fullname}
\bibliography{references}
}

\end{document}